\documentclass[11pt]{article}

\usepackage{geometry}
\usepackage[center]{caption}
\usepackage{makecell}
\usepackage{graphicx}
\usepackage[sort]{natbib}
\usepackage{siunitx}
\usepackage{xspace}
\usepackage{subcaption}
\usepackage{adjustbox}
\usepackage{xr-hyper}
\usepackage{booktabs}
\usepackage{bbding}
\usepackage{enumitem}
\usepackage{multirow}
\usepackage{amsmath}
\usepackage{amssymb}
\usepackage{setspace}
\usepackage{cleveref}
\usepackage{algorithm}
\usepackage{threeparttable}
\usepackage{subcaption}
\usepackage{algorithmic}
\usepackage{array}    
\usepackage{pdfpages}

 \bibpunct[, ]{(}{)}{,}{a}{}{,}%
 \def\bibfont{\small}%
 \def\bibsep{\smallskipamount}%
 \def\bibhang{24pt}%

\def\EquationsNumberedThrough{%
\def\theequation{\arabic{equation}}%
\def\ECEquationsNumberedThrough{%
\setcounter{equation}{0}\let\oldtheequation=\theequation\def\theequation{EC.\oldtheequation}%
}%
\def\ECHowEquations{\ECEquationsNumberedThrough}%
}
\EquationsNumberedThrough    

\newcommand{\nconfigs}{25\xspace}

\newcommand{\companyname}{Bosch\xspace}

\usepackage{mathptmx}
\usepackage[T1]{fontenc}

\usepackage{natbib}
\setcitestyle{aysep={}}

\newenvironment{biseabstract}{%
\begin{quote} \bf}
{\end{quote}}

\newenvironment{bisekeywords}{%
\begin{quote} \it \textbf{Keywords:}}
{\end{quote}}

\title{Predictive maintenance in manufacturing: \linebreak The operational value of continual machine learning} 

\author
{Pascal Janetzky$^{1,2}$, Michael Klar$^{1}$, Manuel Wenzel$^{1}$, Tobias Schlagenhauf$^{2}$, and Stefan Feuerriegel$^{3\ast}$\\
\\
\normalsize{$^{1}$Bosch Center for Artificial Intelligence (BCAI)}\\
\normalsize{$^{2}$Albstadt-Sigmaringen University}\\
\normalsize{$^{3}$Munich Center for Machine Learning (MCML) \& LMU Munich}\\
\\
\normalsize{$^\ast$Corresponding author}
}

\date{}

\begin{document} 

\baselineskip18pt


\maketitle

\begin{biseabstract}
  Predictive maintenance in high-volume manufacturing often faces a \emph{domain-incremental learning problem}: as new product configurations enter production, prediction models must learn from newly available data while preserving performance on earlier configurations that remain in operation. In this paper, we empirically study the operational value of continual machine learning methods for addressing this challenge in predictive maintenance, using real-world maintenance data from \textit{\companyname}. The setting at \textit{\companyname} covers \nconfigs product configurations over five years, which creates a practical challenge for predictive maintenance models, as such models must be updated with data from newly introduced configurations without reducing performance on configurations learned earlier, a phenomenon known as `forgetting'. At the same time, retaining all historical sensor data and repeatedly retraining models is often infeasible in high-volume manufacturing since industrial machines generate too large data volumes. To address this challenge, we combine continual machine learning with robust optimization to support predictive maintenance. We then compare continual machine learning with a broad set of baselines, including fine-tuning, domain-adaptation methods, online learning methods, and configuration-specific models. Across experiments, continual machine learning methods achieve stronger predictive performance and lower maintenance costs than the baselines, with a reduction of cost-related indicators of more than \SI{40}{\percent} relative to the baselines. Our results demonstrate the operational value of continual machine learning for decision support in maintenance when product portfolios grow over time and historical manufacturing data cannot be fully retained.
\end{biseabstract}

\begin{bisekeywords}
Continual machine learning; Decision support system; Domain-incremental learning; Predictive maintenance; Manufacturing
\end{bisekeywords}



\section{Introduction}
Maintaining production machines is crucial for ensuring process quality \citep{ollila1999maintenance}, particularly given that maintenance can account for up to \SI{70}{\percent} of total production costs \citep{madu2000competing}. In the U.S. alone, machine maintenance is estimated to cost \$50 billion annually \citep{thomas2018costs}. To reduce these costs, manufacturers increasingly rely on predictive maintenance, where machine learning models predict maintenance needs before failures occur and thereby support proactive maintenance interventions \citep{thomas2018costs,scaife2024improve}. Predictive maintenance is now widely used across industries such as transportation \citep{rokhforoz2021hierarchical}, consumer goods \citep{mazhar2007remaining}, and manufacturing \citep{menanno2023centrifugal}.

\begin{figure}
    \centering
    \includegraphics[width=\linewidth]{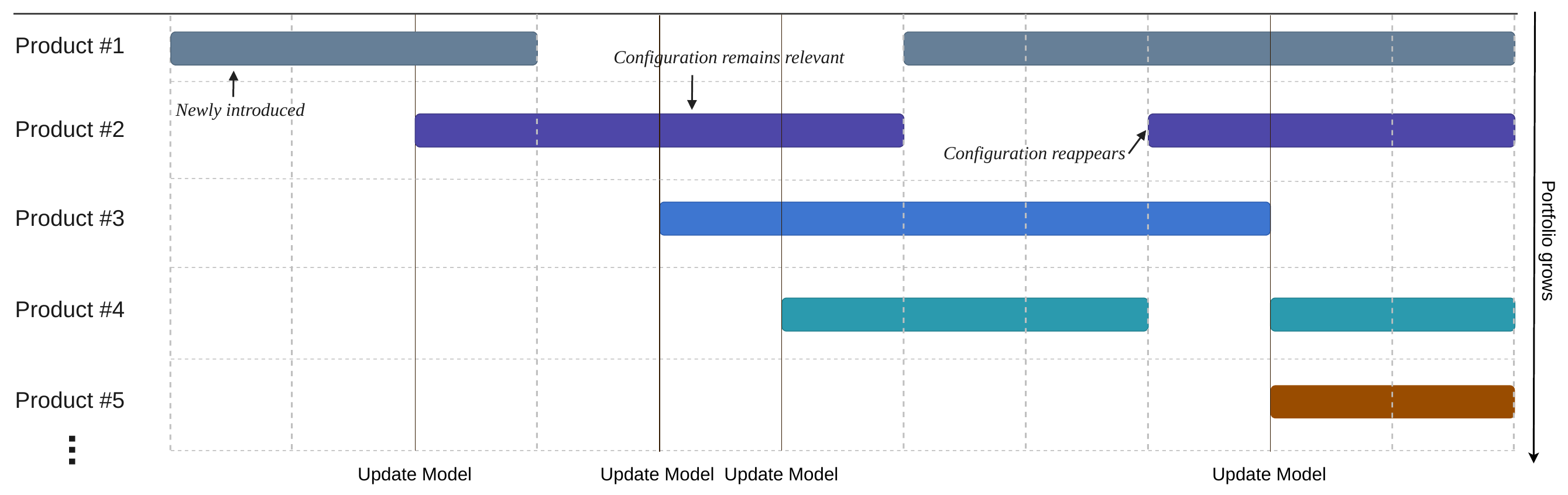}
    \caption{The domain-incremental manufacturing setting at \companyname. Shown is how the product portfolio grows over time while old configurations remain active or reappear. The machine learning models must be regularly updated with new data to work with new configurations and still perform well on old configurations (i.e., to avoid so-called forgetting).}
    \label{fig:figure1}
\vspace{-10pt}
\end{figure}

Predictive maintenance models are regularly updated as new data becomes available to adapt to the new production environment. However, in practice, this task is more difficult than it may initially appear. New data are often generated by newly introduced product configurations that arise from changing customer requirements, new materials, or modifications to production processes. These changes alter machine operating conditions and often lead to different patterns of machine wear and maintenance needs \citep{senoner2023addressing,chen2026ensemble}. At the same time, previously produced configurations may remain in operation or reappear due to recurring orders, seasonal demand, or production planning decisions; see \Cref{fig:figure1} for a conceptual schematic. Consequently, predictive maintenance models must not only learn from newly introduced configurations but also maintain predictive performance on configurations encountered previously. This setting corresponds to a \emph{domain-incremental learning} problem, where new product configurations enter the prediction task while previous configurations remain operationally relevant.

Addressing this problem is challenging for two reasons. First, modern manufacturing systems generate massive volumes of sensor data. For example, a single jet engine can produce more than 1 TB of data per day \citep{jordan2018statistics}, and similar volumes are common in industrial manufacturing environments. Retaining all historical sensor data and repeatedly retraining predictive maintenance models from scratch is therefore often infeasible. Second, na{\"i}vely updating a machine learning model with data from newly introduced product configurations can overwrite knowledge acquired from earlier configurations. As a consequence, predictive performance on previously encountered configurations may deteriorate when these configurations reappear in production. This phenomenon is commonly referred to as \emph{forgetting} or, in machine learning literature, as \emph{catastrophic forgetting} \citep{mccloskey1989catastrophic,zenke2017continual}.

Several machine learning approaches have been proposed for learning from evolving data, yet none directly addresses the requirements of our setting. A common strategy is periodic retraining, where models are re-estimated using both historical and newly collected data \citep{cui2018operational}. However, retraining requires access to historical manufacturing data and is therefore often impractical in high-volume manufacturing environments, where such data cannot be stored. Fine-tuning updates a predictive model using only newly observed data, but it is susceptible to catastrophic forgetting because model parameters adapt exclusively to the most recent observations \citep{zenke2017continual} and will thus have a low performance on older configurations. Similarly, online learning methods are designed to adapt models to incoming observations and evolving distributions \citep{gama2014survey, losing2018incremental}. However, these approaches typically prioritize performance on the current distribution and do \emph{not} explicitly seek to preserve performance on previously observed domains once they are no longer present in the data stream. Finally, manufacturers may maintain separate models for individual product configurations \citep{chen2026ensemble}. While operationally feasible in some settings, such approaches scale poorly as product portfolios expand and do not allow knowledge sharing across configurations.

Motivated by these challenges, we study the operational value of continual machine learning for predictive maintenance in manufacturing. Continual machine learning provides a general strategy for updating machine learning models sequentially while mitigating catastrophic forgetting \citep{kirkpatrick2017overcoming,zenke2017continual}. We empirically analyze the operational value of continual machine learning using real-world maintenance data from \companyname, a leading manufacturing company, covering \nconfigs product configurations over a period of five years. The data are characterized by a domain-incremental learning problem and represent a realistic high-volume manufacturing environment in which historical sensor data cannot be retained for practical reasons and where new product configurations are introduced sequentially over time while old product configurations reappear. For our empirical analysis, we integrate different model-update strategies into a common decision-analytic framework that combines continual machine learning with robust optimization to explicitly account for both domain-incremental learning and predictive uncertainty. We further compare continual machine learning approaches against a broad set of alternatives, including fine-tuning, online learning, and configuration-specific models. 
Across an extensive series of numerical experiments, we find that continual machine learning consistently outperforms these baselines in both predictive performance and maintenance cost and can offer considerable cost savings.

Our paper contributes to research on decision analytics and to managerial practice. From a managerial perspective, we assess the operational value of continual machine learning for predictive maintenance in manufacturing. As manufacturers increasingly face growing product portfolios and large-scale sensor streams, our results show that maintenance models must be updated incrementally but in a way that new information is incorporated without sacrificing performance on earlier configurations (i.e., domain-incremental learning). Failing to do so can lead to suboptimal maintenance decisions and increased operational costs. 
From a research perspective, we contribute to the literature on data-driven decision support \citep{landwehr2022design,meire2026decision} by studying how different model-update strategies affect predictive performance and downstream decisions in domain-incremental settings. While continual machine learning has been studied in computer science {\citep[e.g.,][]{kirkpatrick2017overcoming,janetzky2024slowing}}, we adapt and assess its value for data-driven decision-making under real-world operational constraints in manufacturing environments. Together, our results demonstrate the importance of accounting for evolving product domains when designing and updating machine-learning-based decision systems.

The remainder of our paper is structured as follows. \Cref{sec:rel_work} reviews the current state of predictive maintenance and methods for domain-incremental learning. \Cref{sec:empirical_setup} introduces our empirical setting at \companyname. In \Cref{sec:decision_model}, we present the empirical evaluation framework. \Cref{sec:experimental_setup} describes our experimental setup and benchmark suite. We then report results and operational value of benchmarked methods in \Cref{sec:results}. Finally, in \Cref{sec:discussion}, we discuss the implications of our findings for decision analytics and managerial practice.

\section{Related work}\label{sec:rel_work}
This paper is motivated by the practical problem of updating predictive maintenance models when product portfolios grow over time and historical sensor data cannot be fully retained (``domain-incremental learning''). To situate our work, we review two research streams: predictive maintenance as data-driven decision support and approaches for updating machine learning models under changing product domains. We emphasize how these streams relate to our domain-incremental manufacturing setting, where new product configurations enter production while previously encountered configurations remain operationally relevant or reappear. 

\subsection{Predictive maintenance as data-driven decision support}

Machine learning (ML) is nowadays widely used to support data-driven operational decisions \citep{gupta2023opportunities}. A key application is predictive maintenance, where the risk of machine failures is predicted to enable making proactive maintenance interventions before failures occur \citep[e.g.,][]{mazhar2007remaining,fink2015classification,maschler2021regularization,menanno2023centrifugal}.  Beyond predicting maintenance needs, prior research has also examined how machine learning can support maintenance decision making, such as through deep-learning-enabled decision support systems for infrastructure maintenance \citep{landwehr2022design}. By anticipating failures in advance, predictive maintenance can help reduce downtime and lead to lower overall operational costs \citep{mckone2002guidelines,scaife2024improve}. For a comprehensive review of predictive maintenance, we refer to \citet{jimenez2020towards} and \citet{scaife2024improve}.

Methods for predictive maintenance typically follow a two-step approach based on the predict-then-optimize paradigm \citep[e.g.,][]{bastani2022applied,elmachtoub2022smart,senoner2022using}, which proceeds as follows: 
In the first step, a predictive model is trained to predict the target variable, such as, for example, the probability of machine faults because of tool wear \citep{ayvaz2021predictive}. A wide range of machine learning models have been used for this task, including linear regression, support vector machines, and random forests (see \citet{carvalho2019systematic} for an overview). More recently, neural networks have become the de facto standard in predictive maintenance, particularly due to their ability to learn complex patterns without extensive feature engineering or domain expertise \citep{carvalho2019systematic}. For that reason, our empirical comparison also focuses on neural networks.

In the second step, predictions from the machine learning model are passed into an optimization model that determines the best decisions (e.g., such as the optimal maintenance schedule or replacement decisions) based on a cost function \citep{elmachtoub2022smart}. In general, this optimization step distinguishes predictive maintenance from traditional condition-based maintenance approaches \citep{keizer2018condition,uit2020condition}, which typically focus on assessing the current health state of machinery (e.g., whether a machine is operational or meeting certain production quotas), but not with the aim of directly informing maintenance decisions. In predictive maintenance, the cost function captures the trade-offs between performing maintenance too early (which leads to unnecessary downtime or wasted resources), the direct cost of maintenance, and the potential costs caused by delayed maintenance (such as equipment failures or other downstream disruptions in the production). The effectiveness of this optimization step thus depends heavily on the accuracy of the predictions. While some studies analyze the sensitivity to prediction errors \citep[e.g.,][]{ritzman1993relative,fildes2011incorporating}, the uncertainty inherent in predictive maintenance models is often overlooked. As we show later, ignoring this uncertainty can lead to overly optimistic decisions, especially in domains like manufacturing, where machine-uptime is often highly skewed.

A common challenge is that machine behavior may change as new configurations are introduced. Prior research on predictive maintenance has emphasized the importance of product-specific maintenance strategies \citep{feng2018research}, but often assumes that machine learning models can be re-trained from scratch when needed. In high-volume manufacturing, however, retaining all historical sensor data and repeatedly retraining models from scratch is often infeasible. Moreover, earlier product configurations may remain in production or reappear, so model updates must not come at the expense of predictive performance on previously encountered configurations. This creates a so-called domain-incremental learning problem: predictive maintenance models must incorporate newly available data from different product domains while preserving performance on earlier domains (cf. \Cref{fig:figure1}). We next position this setting relative to adjacent streams on updating machine learning models under changing product domains.

\subsection{Updating machine learning models under changing product domains}

Several literature streams study how prediction models should be updated when new data become available over time and how changes in the underlying data distributions necessitate model updates for data-driven decision support \citep{gama2014survey,rolnick2019experience,shalev2012online,meire2026decision}. Existing approaches differ in the assumptions they make about how often changes in the product domains occur, whether earlier domains remain relevant, and whether historical data can be stored (see \Cref{tab:related_work} and \Cref{sec:extended_related_work} in the Online Appendix). These dimensions characterize the domain-incremental learning setting in our paper and show how it is different from other settings: product configurations are added sequentially, previously encountered configurations may remain in production or reappear, and historical sensor data cannot be fully stored. The key challenge in domain-incremental learning is therefore not only to adapt predictive maintenance models to newly introduced configurations, but also to preserve performance on previously encountered configurations that remain operationally relevant without storing the full dataset. We discuss the differences to other settings below. 

\begin{table}[htbp]
\centering
\tiny
\caption{Comparison of different approaches for updating machine learning models under growing product portfolios in manufacturing.}
\label{tab:related_work}
\begin{threeparttable}
\begin{tabular}{l|m{3.0cm}|c|c|>{\arraybackslash}m{4.5cm}}
\toprule
Research stream & High-level objective & \makecell[c]{New configurations\\added repeatedly?} & \makecell[c]{Old configurations\\remain/reappear?} & Application \\
\midrule

\shortstack[l]{\textit{Na\"ive approaches}}
& Repeatedly fine-tune a model; train domain-specific models
& \CheckmarkBold
& \XSolid
& e.g., \citet{bastani2021predicting,ng2023augmenting,niu2026internal} \\

\midrule

\shortstack[l]{Domain adaptation}
& Adapt a model from a source domain to a target domain
& \CheckmarkBold
& \XSolid
& e.g., \citet{bastani2021predicting,ng2023augmenting,senoner2023addressing}\\

\midrule

\shortstack[l]{Online learning}
& Adapt to incoming data or a changing distribution
& \CheckmarkBold
& \XSolid
& e.g., \citet{zenisek2019machine,chen2023network,ferreira2023demand}\\

\midrule

\shortstack[l]{Domain-incremental learning}
& Learn new domains while retaining performance on earlier domains; constraint: data cannot be stored due to volume
& \CheckmarkBold
& \CheckmarkBold
& This paper \\

\bottomrule
\end{tabular}
\end{threeparttable}
\vspace{-10pt}
\end{table}

\emph{Na\"ive approaches} serve as simple benchmarks for updating predictive models under changing product domains. Examples include static models, which are trained once and then kept fixed~\citep{carvalho2019systematic,losing2018incremental}; fine-tuning, which updates the model on newly available data~\citep{bastani2021predicting}; and per-configuration models, which train separate models for individual product configurations~\citep{bjarnadottir2020machine}. Because these approaches are usually straightforward to implement, they often can be a first choice in practical settings. However, they address only parts of our domain-incremental setting. Static models do not learn newly introduced configurations, fine-tuning can overwrite knowledge about earlier configurations and thus lead to forgetting, and per-configuration models do not share information across product domains and scale poorly as the product portfolio grows. Hence, these approaches provide useful baselines, but they do not directly address the need to learn new configurations while retaining performance on earlier ones.

\emph{Domain adaptation} is focused on how a model can be transferred from a source domain to a target domain \citep{kouw2018introduction,kouw2019review}. Typical domain adaptation approaches include data reweighting, aligning feature representations, and parameter-transfer methods \citep{loog2012nearest,wang2018deep}. Commonly, domain adaptation assumes a limited, often one-time, shift between source and target domains, where the main objective is to achieve a strong performance on the target domain after adaptation, while little emphasis is placed on retaining performance on the source domain. From a manufacturing perspective, domain adaptation is useful when predictive models must be transferred across related but different operational settings, such as different products, machines, or plants. Several prior works have therefore examined the operational value of domain adaptation in such setting \citep[e.g.,][]{bastani2021predicting,ng2023augmenting,senoner2023addressing}. For example, \citet{senoner2023addressing} study distributional shifts in customized production and show how prediction models can be adapted to improve throughput-time predictions for new customer orders with different specifications. However, domain adaptation is not suited for domain-incremental learning settings, where new product domains accumulate over time but where earlier configurations additionally remain relevant.

\emph{Online learning} focuses on sequential updates under computational or storage constraints \citep{losing2018incremental,shalev2012online,zenisek2019machine,chen2023network,ferreira2023demand}. A specific focus is drift learning, which considers settings in which the data-generating process changes gradually over time (e.g., when machine conditions or process parameters change during ongoing production due to tool wear). In many such settings, previously learned concepts become outdated and are therefore discounted or replaced to maintain the quality of data-driven decisions \citep[cf.][]{gama2014survey, lu2020data}. Common online learning approaches include sliding-window, adaptive ensemble, and streaming-tree methods \citep{oza2001online,bifet2009new,gama2014survey}. These methods are relevant for data-driven decision support when models must be adapted quickly to changing data. For example, prior studies have used online or drift-aware learning to update models in settings such as manufacturing, network operations, and demand forecasting \citep{zenisek2019machine,chen2023network,ferreira2023demand}. However, online learning methods are not useful in our setting because online learning typically prioritizes performance on the new configurations and does \emph{not} explicitly seek to preserve performance on previously encountered configurations that remain operationally relevant or reappear.

\emph{Domain-incremental learning} describes settings in which new domains are added over time while earlier domains remain relevant or reappear. In our manufacturing setting, this means that predictive maintenance models must learn from newly introduced product configurations without losing performance on configurations encountered earlier. This also characterizes the manufacturing data at our partner company (see \Cref{tab:dataset_summary}). The key challenge is forgetting, also known as catastrophic forgetting, where updating a model on new domains can overwrite knowledge learned from earlier domains \citep{kirkpatrick2017overcoming,rolnick2019experience,zenke2017continual}. Another key challenge is that high-volume manufacturing generates large amounts of raw sensor data, so model updates should \emph{not} require that the full historical dataset is stored. Continual machine learning is explicitly designed for the objective in domain-incremental learning; that is, learning on new domains while retaining performance on earlier ones \citep{kirkpatrick2017overcoming,zenke2017continual}. 

A prominent approach is progressive neural networks (PNN) \citep{rusu2016progressive}. PNN mitigates forgetting by keeping model components learned for earlier domains fixed and adding new domain-specific components when a new domain is introduced. As we will show later, this approach is particularly useful in manufacturing settings with domain-incremental learning because it offers a principled mechanism to update the neural network architecture by adding domain-specific model components, so that it directly accommodates new product configurations while preserving those learned for earlier configurations. We provide a detailed overview of continual machine learning methods in Online Appendix~\ref{sec:extended_related_work}. However, the operational value of continual machine learning in manufacturing remains unclear, especially when predictions are translated into maintenance decisions and downstream costs. Existing studies on continual machine learning in manufacturing often rely on simulated settings or address tasks other than predictive maintenance~\citep{maschler2021regularization,tercan2022continual}. To fill this gap, we empirically evaluate the operational value of  continual machine learning for supporting maintenance decisions in a domain-incremental manufacturing setting.

\section{Empirical setup at \companyname}\label{sec:empirical_setup}
\subsection{Manufacturing at \companyname}
Our empirical application for this study is conducted at \companyname, headquartered in Gerlingen, Germany. Founded in 1886, \companyname is a global leader in engineering and technology, and operates across various sectors, including mobility, industrial technology, consumer goods, and energy and building technology. With over 400,000 employees worldwide, \companyname invests heavily in research and development to maintain its reputation for high product quality and technological innovation.

We focus on a high-volume manufacturing process that produces metallic components as parts of steering units for the automotive sector. The specifications of the components vary due to different customer requirements. This is the case, for example, because customers introduce a new car model based on a different design, which thus requires a different product configuration. At the same time, customers may continue to reorder older parts, so previously produced configurations can remain in production or reappear later. As a result, the product portfolio grows over time: manufacturers must accommodate newly requested configurations while still maintaining production capabilities for earlier ones.

To maintain consistent product quality, maintenance is regularly conducted on the precision machining tools and, when necessary, worn components inside the machine are replaced. For instance, switching from a softer material like aluminum to a harder one, such as high-strength steel, can greatly accelerate tool degradation. However, maintenance activities result in production downtime and possible revenue loss. Delaying maintenance, on the other hand, can be even more costly due to defects, downstream disruptions, or even rework at later stages in the manufacturing process. Currently, tool wear is assessed manually at fixed intervals. If a manual inspection reveals signs of tool degradation, the tool is replaced. However, these manual assessments incur high costs: (i)~labor and time costs, and (ii) opportunity costs due to production halts. To optimize the inspection and maintenance process, \companyname partnered with us to empirically compare alternative model-update strategies for predictive maintenance.

\subsection{Operational setting}
For this study, a subset of the production machines was manually equipped with recording sensors to capture machine data. The data collection was limited to three machines, as continuous recording across all machines or even the entire facility is not feasible due to the immense volume of data generated in high-throughput manufacturing (i.e., up to several hundreds of gigabytes per machine per day), so storing the entire generated data is practically impossible; instead, only a subset of the generated production data can be temporarily retained. Thus, at fixed intervals, the collected data was manually transferred from the machines to a central system. Moreover, for safety reasons, manufacturing machines are often located in access-controlled areas, because of which real-time external access is often limited. 

For the purpose of this work, we received data from \nconfigs product configurations, including detailed information about machine health (i.e., tool degradation), covering a period of five years. Over this period, new product configurations were introduced while earlier ones were sometimes suspended but also reappeared. As such, the empirical application at \companyname is characterized by a \emph{domain-incremental learning} problem. Throughout this paper, we refer to a \emph{product configuration} as a unique combination of a specific product specification (e.g., the length of a component) and the associated parameter setting on the production machines (e.g., amount of cooling liquid necessary).

To characterize the domain-incremental nature of our setting, we illustrate the timeline of product configurations in \Cref{tab:dataset_summary}. We observe that, over time, product configurations changed at \companyname, with new configurations being introduced, while earlier ones often remain relevant and some reappear. For example, late in year 1, the machines were primarily used for three main product configurations, while we see a notable surge in market demand for other product configurations after year 3, resulting in increased product variety over time (overall \nconfigs product configurations across the full observation period). This recurring pattern makes maintaining a high performance on previously encountered configurations operationally important. 

\begin{table}[tbp]
\tiny
    \centering
    \caption{Timeline of product configurations in our given manufacturing data, which are sometimes temporarily suspended and may later reappear. Here, product configurations refer to unique product specifications and the corresponding machine settings. Reported is the number of samples for each product configuration in each quarter.}
    \label{tab:dataset_summary}

\begin{tabular}{l|r rr rrrr rrrr r|r}
\toprule
Product & Year 1 & \multicolumn{2}{c}{Year 2} & \multicolumn{4}{c}{Year 3} & \multicolumn{4}{c}{Year 4} & Year 5 &  \\
\cmidrule(lr){2-2} \cmidrule(lr){3-4} \cmidrule(lr){5-8} \cmidrule(lr){9-12} \cmidrule(lr){13-13}
configuration & Q4 & Q1 & Q2 & Q1 & Q2 & Q3 & Q4 & Q1 & Q2 & Q3 & Q4 & Q1 & \#Samples \\
\midrule
\#1 & 2516 & 11395 & 2226 & --- & --- & --- & --- & --- & --- & --- & --- & --- & 16137 \\
\#2 & 8399 & 10424 & 4189 & --- & --- & --- & --- & --- & --- & --- & --- & --- & 23012 \\
\#3 & 2121 & 3906 & 1512 & --- & --- & --- & --- & --- & --- & --- & --- & --- & 7539 \\
\#4 & --- & --- & --- & 1470 & --- & 3669 & 1429 & 9757 & 7993 & 2580 & --- & --- & 26898 \\
\hline
\#5 & --- & --- & --- & 3918 & 10772 & 17236 & --- & 5932 & --- & --- & --- & --- & 37858 \\
\#6 & --- & --- & --- & 1959 & 1083 & 4148 & --- & --- & --- & --- & --- & --- & 7190 \\
\#7 & --- & --- & --- & 6758 & 4452 & 3969 & --- & 282 & 1382 & 177 & --- & --- & 17020 \\
\#8 & --- & --- & --- & --- & 2069 & 3529 & 1362 & 3772 & --- & --- & --- & --- & 10732 \\
\#9 & --- & --- & --- & --- & 1256 & 253 & --- & 1660 & --- & --- & --- & --- & 3169 \\
\#10 & --- & --- & --- & --- & 475 & 484 & --- & 694 & --- & --- & --- & --- & 1653 \\
\hline
\#11 & --- & --- & --- & --- & 934 & 1915 & --- & 1177 & 1542 & --- & --- & --- & 5568 \\
\#12 & --- & --- & --- & --- & 1113 & 1178 & 2077 & --- & --- & --- & --- & --- & 4368 \\
\#13 & --- & --- & --- & --- & 2559 & 1070 & 1011 & 296 & --- & --- & --- & --- & 4936 \\
\#14 & --- & --- & --- & --- & 414 & --- & --- & --- & --- & --- & --- & --- & 414 \\
\#15 & --- & --- & --- & --- & --- & 1963 & 2474 & --- & --- & --- & --- & --- & 4437 \\
\hline
\#16 & --- & --- & --- & --- & --- & 2045 & --- & --- & --- & --- & --- & --- & 2045 \\
\#17 & --- & --- & --- & --- & --- & 875 & --- & --- & --- & 1525 & 1866 & 1420 & 5686 \\
\#18 & --- & --- & --- & --- & --- & 278 & --- & 3543 & 3394 & 5880 & 6445 & --- & 19540 \\
\#19 & --- & --- & --- & --- & --- & --- & 611 & 4354 & 2801 & --- & --- & --- & 7766 \\
\hline
\#20 & --- & --- & --- & --- & --- & --- & --- & 3103 & 5552 & 465 & 5590 & --- & 14710 \\
\#21 & --- & --- & --- & --- & --- & --- & --- & 4313 & --- & --- & --- & --- & 4313 \\
\#22 & --- & --- & --- & --- & --- & --- & --- & --- & 319 & --- & 757 & --- & 1076 \\
\#23 & --- & --- & --- & --- & --- & --- & --- & --- & 4028 & --- & --- & 2036 & 6064 \\
\#24 & --- & --- & --- & --- & --- & --- & --- & --- & 789 & --- & --- & 1240 & 2029 \\
\#25 & --- & --- & --- & --- & --- & --- & --- & --- & --- & 1232 & 732 & 449 & 2413 \\
\bottomrule
\multicolumn{14}{l}{{Note: the machines were temporarily offline in Q3/Q4 in Year 2.}}
\end{tabular}
\vspace{-10pt}
\end{table}

\subsection{Empirical task}
We were tasked with the problem of optimizing maintenance decisions in the above manufacturing setting, which is characterized by domain-incremental learning, where new product domains are introduced over time while earlier product configurations remain operationally relevant. The resulting task is to learn from newly introduced domains without reducing performance on previously encountered product configurations. Here, a key challenge is \emph{forgetting} \citep[cf.][]{kirkpatrick2017overcoming,wang2024comprehensive}, which occurs when new information overwrites previously learned knowledge about earlier product configurations. 

One may think that several straightforward update strategies could address this task, but each has important limitations in our setting. The first is full re-training \citep{cui2018operational}; that is, training a new model from scratch on the combined dataset with historical and new data. However, this is practically infeasible: precision machining can produce several gigabytes or even petabytes of manufacturing data per year. In practice, retraining a model with such large-scale data is \emph{not} possible. 
The second option is to train separate, configuration-specific models. This can be attractive when known configurations recur, but it scales with the number of configurations and does \emph{not} exploit potential information sharing across them. Similarly, fine-tuning, domain adaptation and online learning provide useful tools for adapting models to changing data, but they do \emph{not} explicitly preserve the performance on earlier configurations that may remain relevant or reappear. We therefore focus on continual machine learning, which is explicitly designed for domain-incremental learning.

\subsection{Manufacturing data}
To empirically evaluate alternative model updating strategies for predictive maintenance, we received access to a real-world dataset from our partner company. The dataset includes maintenance events for $P = \{1, \ldots, \nconfigs\}$ product configurations observed over time. Due to the immense volume of data generated in high-throughput manufacturing, only a subset of the generated production data could be retained for this study. The data collection was designed by the company, so that the data accurately reflect the properties of machine behavior in real-world production. In total, we have \num{236573} observations in the sensor data for the study period.

Each observation $\mathbf{x} \in \mathrm{R}^{d_x}$ is associated with a binary label $y \in \{0, 1\}$ that represents the maintenance decisions. At \companyname, maintenance decisions for our manufacturing process are made as follows: machine states are evaluated by multiple domain experts, and machining tools are replaced when the domain experts determined that these are worn out (i.e., $y=1$). Overall, due to a conservative and high-quality-focused maintenance strategy at the partner company that prioritizes reliability and thus involve frequent manual inspections, tool replacements are unevenly distributed, with approximately \SI{90}{\percent} of the data falling into the \emph{no tool replacement} category (i.e., $y=0$). We carefully address later the class imbalance during model training; details are provided in \Cref{sec:experimental_setup}.

Each observation $\mathbf{x}$ consists of $d_x=932$ sensor-based features. These features are captured via three different high-frequency sensors mounted on the machine. These sensors capture different industry-standard measurements (e.g., pressure, temperature, torque). The raw time-series signals are then automatically pre-processed into an array of  \num{932} scalar values that represent statistical characteristics of the original signals. For this step, our partner company employs widely used time-series processing techniques. Examples of such methods are the Fourier transformation, power spectral density at various frequencies, and normalized energy from wavelet transformation. From all transformation methods, aggregate statistics are computed at different frequencies. A description of the final features is provided in \Cref{tab:feature_list}. Due to confidentiality constraints, we report only a summarized description of the feature groups rather than the complete list of individual sensor features.

\begin{table}[htbp]
\centering
\footnotesize
\caption{List of features captured during manufacturing, recorded by sensors monitoring the production process and used as input to the predictive maintenance model.}
\label{tab:feature_list}
\begin{tabular}{lll}
    \toprule
    Feature range & Description & Example \\
    \midrule
    1--394   & Sensor 1 features & Normalized energy of wavelet packet transform \\
    395--663 & Sensor 2 features & Mean frequency of Fourier transform \\
    664--932 & Sensor 3 features & Median power spectral density at various frequencies \\
    \bottomrule
\end{tabular}
\vspace{-10pt}
\end{table}

In addition to sensor-based features, each observation is accompanied by \textit{meta-information} $\mathbf{m} \in \mathbb{R}^{d_m}$, containing information about product configurations such as material type, material supplier, and previous manufacturing steps, among others. In our dataset, we have $d_m=40$ values with meta-information, that contain details about material type, material supplier, and previous manufacturing steps, among other factors. A list of the meta-information is given in \Cref{tab:meta_information_list}.

\begin{table}[htbp]
\centering
\footnotesize
\caption{List of meta-information, which is further grouped into (i)~\emph{machine features} that describe the machine and (ii)~\emph{product features} that describe the product.}
\label{tab:meta_information_list}
\begin{tabular}{lll}
    \toprule
    Variable range & Description & Examples \\
    \midrule
    1--17  & Machine features & Machine type, part counter, timestamp \\
    18--40 & Product features & Length, material type, supplier \\
    \bottomrule
\end{tabular}
\vspace{-10pt}
\end{table}

Because machine states are inspected only at irregular intervals, some labels are missing between consecutive assessments. We thus back-fill missing labels between two identical assessments to avoid information leakage. We z-score normalize all continuous input features using statistics computed on the training data and apply the same transformation to the corresponding test data. We encode categorical meta-information, such as supplier and material type, using one-hot encoding. In the main experiments, we use only sensor-based features; we study the additional predictive value of meta-information in an ablation analysis.

\subsection{Exploratory data analysis}\label{sec:exploratory}

In this section, we explore how the data differ across product configurations over time. Inspired by \citet{senoner2023addressing}, we employ three methods for this purpose: (1)~pairwise Kullback-Leibler (KL) divergence, (2)~comparing a single-configuration model with an all-configuration model, and (3)~a Kolmogorov-Smirnov (KS) test to assess the difference in feature values over time.

\begin{figure}[!htbp]
\begin{minipage}[r]{0.48\linewidth}
\centering
\includegraphics[width=0.7\linewidth]{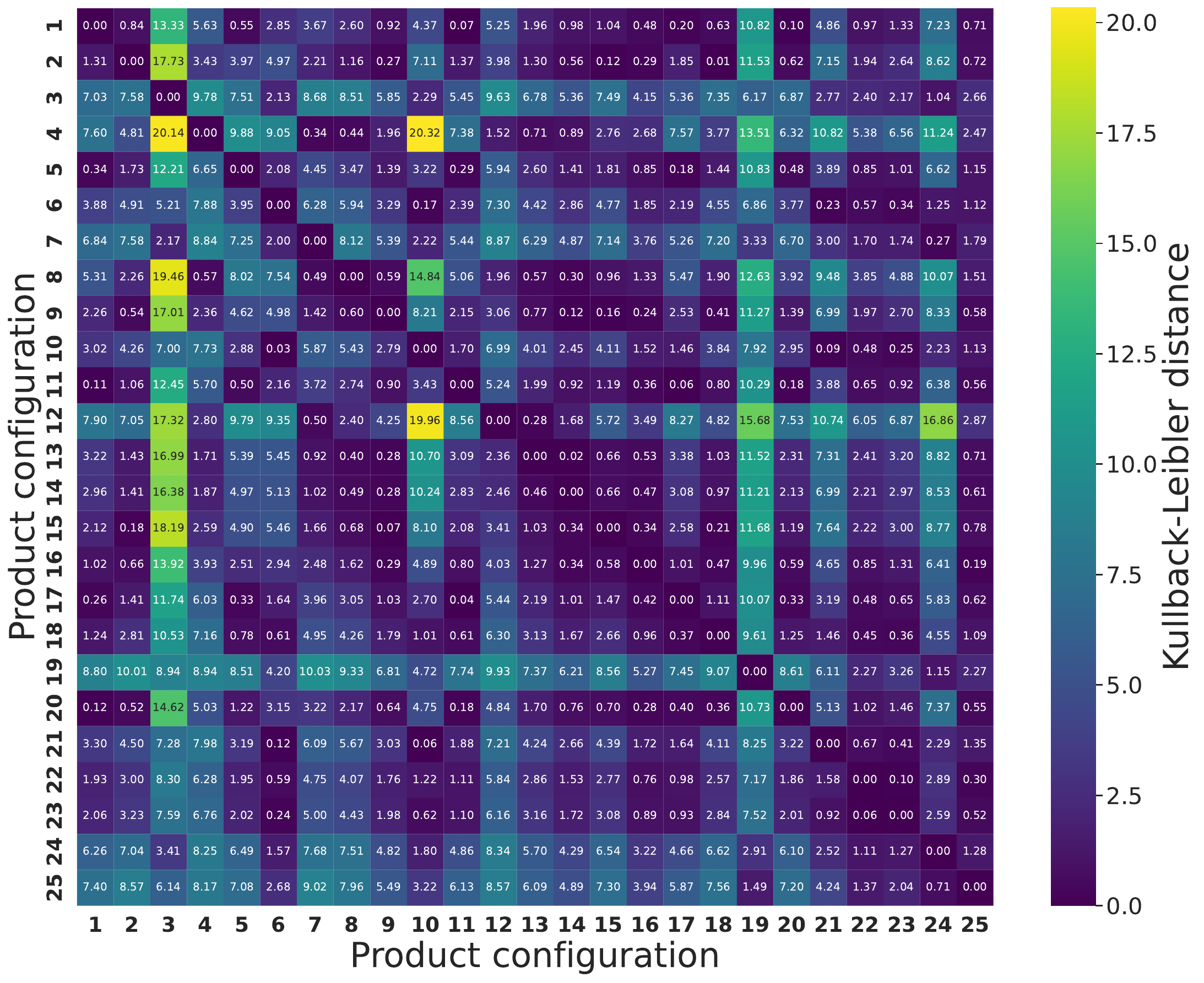}
     \caption{Heatmap showing pairwise Kullback-Leibler (KL) divergences between the data of product configurations. Note that KL divergence is not symmetric, so values may differ depending on the direction of comparison.}
     \label{fig:kullback_plot}
\end{minipage}
\hfill
\begin{minipage}[c]{0.48\linewidth}
\includegraphics[width=\linewidth]{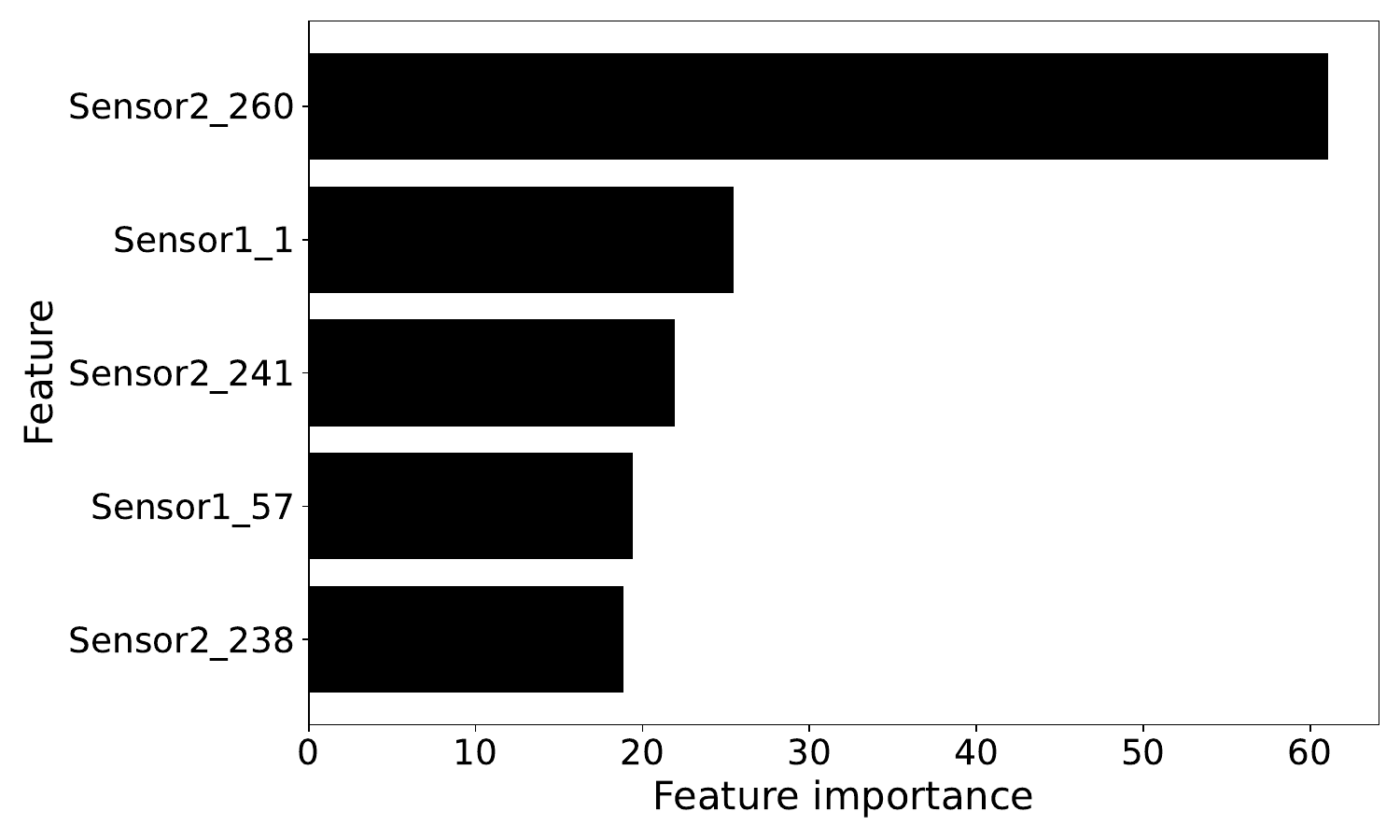}
     \caption{The five most important features for explaining the difference between product configurations over time. Evidently, changes in recordings from sensor 2 are strong determinants.}
     \label{fig:top_features_plot}
\end{minipage}
\vspace{-5pt}
\end{figure}

First, to quantify distributional differences between product configurations, we compute the KL divergence between the features of different product configurations. The KL divergence measures how one probability distribution diverges from a second, reference distribution; a KL score of 0 indicates that the two distributions are identical, while higher values indicate a larger difference. \Cref{fig:kullback_plot} visualizes the pairwise KL scores and confirms that there is substantial variation in the configurations, particularly when comparing earlier and later products. This observation shows that product configurations differ substantially in their feature distributions.

Second, we examine how well a model trained on one product configuration performs when applied to other configurations. For this purpose, we compare such model against a full-data model that is trained jointly on data from all configurations. Classification performance is measured using the F1 score, where a higher value indicates better identification of maintenance needs across configurations. We implement this analysis using neural networks (details in \Cref{sec:experimental_setup}) and repeat it ten times. Across all repetitions, we find a large performance gap between the model trained on a single configuration (F1: 0.52) and the jointly trained full-data model (F1: 0.92). This result suggests that predictive patterns differ substantially across product configurations, while also indicating that knowledge sharing across configurations can improve predictive performance. 


Third, we examine whether the changes in features across configurations are statistically significant. To do this, we identify the five most important features using adversarial validation \citep{ng2023augmenting} and then train a multi-class classifier to predict the product configuration from which an observation was drawn, using the unique configuration identifier stored in the meta-information as labels. We visualize the five most important features in \Cref{fig:top_features_plot}. We perform a KS test to compare features for a configuration from year 1 with those from a configuration from year 5. For all features, the differences between the configurations are statistically significant ($p<0.05$), which further confirms that the manufacturing data change over time. 

In summary, our exploratory analysis reveals that the data associated with different product configurations differ substantially. When predictive maintenance models are only updated towards newly-introduced configurations, their performance will be suboptimal on previously seen configurations that remain in production or reappear.

\section{Empirical framework}\label{sec:decision_model}
\subsection{Problem and operational constraints}

We consider a predictive maintenance setting, where machine states are predicted from sensor data and where maintenance decisions are subsequently selected to minimize the expected maintenance cost. In our setting, we consider multiple independent machines (which is common in manufacturing operations at \companyname and other manufacturers), and, for ease of notation, we thus omit machine-specific indices. Let $P=\{1, \ldots ,n_P\}$ denote the set of product configurations observed in the study (here: $n_P = \nconfigs$). From a learning perspective, each product configuration corresponds to a product domain. These domains become available sequentially over time, while previously observed domains may remain relevant or reappear (i.e., domain-incremental learning). 

For each product domain $p \in P$, we observe a set of sensor measurements $\mathbf{x}_{p,i} \in \mathcal{X} = \mathbb{R}^{d_x}$, meta-information $\mathbf{m}_{p,i}\in \mathcal{M} = \mathbb{R}^{d_m}$, and maintenance labels $y_{p,i}\in \{0,1\}$, where $i \in \mathcal{I}_p$ indexes observations within domain $p$. The sensor data capture machine behavior, while the meta-information encode additional information about the process (e.g., material type, machine settings, cooling liquid) that are not otherwise captured in the sensor data. Because manufacturers cannot retain all historical sensor data, predictive models must be updated sequentially as new product domains become available, without access to the full history of past observations.

Let $y_{p,i}$ denote the machine state (or label), where a value of $1$ denotes  ``machine requires maintenance'' and $y_{p,i}=0$ denotes ``machine does not require maintenance''. However, since the true machine state is a~priori unknown, it needs to be predicted from the sensor data. We denote the predicted machine state by $\hat{y}_{p,i} \in [0,1]$, which gives the predicted probability of a maintenance need. For this, we later learn a predictive model $h$ that estimates the machine state $\hat{y}_{p,i} = h(\mathbf{x}_{p,i}, \mathbf{m}_{p,i})$. However, due to the data storage constraints, the predictive models must be updated over time under limited data retention, using newly available data from the current product domain; that is, using $(\mathbf{x}_{p,i}, \mathbf{m}_{p,i})$ but without storing large amounts of historical data.

The maintenance costs are computed as follows. The predicted probabilities of maintenance needs, $\hat{y}_{p,i}$, are mapped onto a (binary) maintenance decision ${d}_{p,i} \in \{0, 1\}$ via thresholding. Formally, we use a prediction threshold $\beta \in [0,1]$, so that
\begin{equation}
    d_{p,i} = \begin{cases}
1, & \text{if } \hat{y}_{p,i} > \beta, \\
0, & \text{otherwise} ,
\end{cases}
\end{equation}
where $d_{p,i}=1$ indicates a decision to perform maintenance, and $d_{p,i}=0$ indicates no maintenance. For this, the cost function $\mathcal{C}$ captures (i)~the costs $c_\mathrm{fp}$ due to type-I errors (maintenance is conducted too early), (ii)~the costs $c_\mathrm{fn}$ due to type-II errors (maintenance is conducted too late, which incurs downstream costs), and (iii)~the actual maintenance costs $c_{\mathrm{m}}$ due to the maintenance event. Generally, the costs of (iii) are substantially lower than (ii) and (i), which implies that preemptive maintenance can be financially cost-effective. The maintenance cost for observation $i$ in domain $p$ is then given by:
\begin{equation}
    \mathcal{C}(y_{p,i}, d_{p,i}) = c_{\mathrm{fp}} \cdot \mathbb{I}(y_{p,i} = 0 \wedge d_{p,i} = 1) + c_{\mathrm{fn}} \cdot \mathbb{I}(y_{p,i} = 1 \wedge d_{p,i} = 0) + c_{\mathrm{m}}\cdot\mathbb{I}(y_{p,i} = 1 \wedge d_{p,i} = 1) ,\label{eq:cost_individual}
\end{equation}
which depends on the maintenance decision $d_{p,i}$ (and thus how the predictions are translated into decision-making) as well as the true but unobservable machine state $y_{p,i}$. Here, $\mathbb{I}$ is the indicator function assessing whether maintenance ($d_{p,i} = 1$) is conducted at a time point where the machine needs maintenance (${y}_{p,i}= 1$).\footnote{For ease of notation, we assume that the maintenance costs are similar across all product configurations. However, accounting for heterogeneity in the maintenance cost across product configurations is straightforward by introducing configuration-dependent costs $c_{\mathrm{fp}}$, $c_{\mathrm{fn}}$, and $c_{\mathrm{m}}$.} The total maintenance cost across all domains and observations is then given by
\begin{equation}
     \mathit{COST} = \sum_{p} \sum_i \mathcal{C}(y_{p,i}, d_{p,i}).\label{eq:costs}
\end{equation}

\subsection{Decision model}

Our decision model follows a two-step, predict-then-optimize approach. In the first step, we learn a predictive machine learning model $h: \mathcal{X}\times \mathcal{M}\rightarrow [0, 1]$ to predict the probability of maintenance need $y_{p,i}$ from sensor data $\mathbf{x}_{p,i}$ and meta-information $\mathbf{m}_{p,i}$. In the second step, we determine the cost-effective maintenance decisions $d_{p,i}$ by optimizing the maintenance cost from \Cref{eq:costs}.

Such a predict-then-optimize approach is widely used  to support data-driven operational decision making \citep[e.g.,][]{bastani2022applied,feng2023framework,jakubik2022data,senoner2023addressing} and is also well suited for our empirical application for three reasons. First, it allows managers to evaluate the trained prediction model to ensure it meets internal performance standards. Second, it provides flexibility in the choice of prediction models, so that managers can select machine learning models that are best suited to their specific production environment, including customized neural network architectures. Third, this approach allows to model predictive uncertainty in the first step and then incorporate the predictive uncertainty in the optimization step through robust optimization. This is especially important in our setting, where machine failures are rare and costly, so overconfident but incorrect predictions can lead to large downstream costs, while the robust optimization can account for how confident the machine learning model is to better balance operational risks.

\subsubsection{Step~1: Prediction model}

We update the predictive machine learning model $h$ sequentially as new product domains become available. Different from traditional predictive maintenance settings with a single fixed model \citep[e.g.,][]{fink2015classification,menanno2023centrifugal}, our setting learns a sequence of models over time. We denote the model after observing product domain $p$ by $h_p(\theta_p,\cdot)$, where $\theta_p \in \Theta$ are the corresponding model parameters. The parameters are updated when new product domains become available using a continual machine learning approach (as described later).

When training $h_{p}$, we aim to minimize the prediction error between the predicted machine states $\hat{y}_{p,i}$ and the true machine states $y_{p,i}$ for observations $i \in \mathcal{I}_p$ in domain $p$. Hence, the objective for updating the model for a new domain $p$ is given
\begin{equation}
\underset{\theta_p \in \Theta}{\mathrm{arg\,min}}\; \sum_{i\in \mathcal{I}_p} \mathcal{L}\big(h_p(\theta_p, \mathbf{x}_{p,i},  \mathbf{m}_{p,i}), y_{p,i}\big),\label{eq:main}
\end{equation}
where $\mathcal{L}$ is a loss function, typically a cross-entropy loss, to measure the error between a model prediction $h_p(\theta_p, \mathbf{x}_{p,i}, \mathbf{m}_{p,i})$ and ground-truth $y_{p,i}$. 

We implement $h(\theta,\cdot)$ as a neural network for three practical advantages: First, neural networks are increasingly used in predictive maintenance scenarios and have been shown to empirically work well \citep[e.g.,][]{carvalho2019systematic,guo2023transformer,zonta2022predictive}. Second, the manufacturing data often contains nonlinear dependencies between product types and their effect on machine states \citep[e.g.,][]{senoner2022using}. Neural networks can learn such complex relationships due to the universal approximation theorem \citep{hornik1989multilayer}. Third, neural networks make use of gradient-based optimization, which is often needed for state-of-the-art continual learning methods.

Finally, in manufacturing practice, an ensemble of predictive models is typically used to improve performance \citep{song2024uncertainty,srinivasan2023ensemble}. As a result, we not only obtain a point prediction but capture the uncertainty around the estimated mean maintenance need $\overline{y}_{p,i}$, which we denote by an interval $\tilde{y}_{p,i} \in [\overline{y}_{p,i} - \delta_{p,i},\; \overline{y}_{p,i} + \delta_{p,i}]$. 
Both $\overline{y}_{p,i}$ and $\delta_{p,i}$ are obtained from the ensemble, which is a standard approach for capturing predictive uncertainty. In our implementation, we use an ensemble of 10 independently trained models for each method instantiation and set the $\delta$ to one standard deviation.

To mitigate the large class imbalances in our partner company data, with approximately 90\% of observations belonging to the “no-maintenance” class, we employ resampling of the data. Specifically, during training, we follow established practice and apply under-sampling to the majority class \citep{krawczyk2016learning,de2023systematic}, while leaving the test distribution unchanged so that evaluation reflects the observed class ratio in the real-world data. We provide further implementation details (e.g., regarding the solver and hyperparameter tuning) in Online~Appendix \ref{sec:implementation_details} and \ref{sec:hparam_tuning_details}.

\subsubsection{Step~2: Robust optimization}
We now translate the predictive model from Step~1 into cost-effective decisions. Here, we account for the predictive uncertainty and thus adopt concepts from robust optimization \citep{bertsimas2004price,bertsimas2011theory}. Specifically, we follow the $\Gamma$-robust framework introduced by \citet{bertsimas2004price}, which provides a principled way to balance performance and robustness. For this, we use the mean maintenance need $\overline{y}_{p,i}$ and uncertainty bound $\delta_{p,i}$ from an ensemble over multiple prediction models used in Step~1. To include the uncertainty bound in the thresholding approach from above, we proceed by utilizing $\delta_{p,i}$ to compute a $\Gamma$-robust decision via
\begin{equation}
    d^{(r)}_{p,i} = \begin{cases}
1, & \text{if } \overline{y}_{p,i} + z_{p,i} \delta_{p,i} > \beta, \\
0, & \text{otherwise}. \label{eq:decision_robust}
\end{cases}
\end{equation}
where $z_{p,i} \in \pmb{z}$, and $\pmb{z}\in \mathcal{Z}= \left\{\pmb{z}: \|\pmb{z}\|_{\infty} \leq 1, \|\pmb{z}\|_1 \leq \Gamma\right\}$ is an uncertainty set that controls the number of predictions that are allowed to simultaneously deviate maximally (i.e., $ z_{p,i} \delta_{p,i}$ with $z_{p,i} = 1$) from their mean, with $\Gamma$ as the upper bound.

Then, we can formulate the robust optimization problem as
\begin{align}
    &\underset{\beta}{\text{min}}\; \underset{\pmb{z}\in \mathcal{Z}}{\text{max}} \sum_p \sum_i \mathcal{C}(y_{p,i}, d^{(r)}_{p,i})\\
    \text{s.t.} &\;\; \beta \in [0,1]  ,
\end{align}
where the robustness parameter $\Gamma$ controls the strength of the uncertainty considered in the optimization process. Here, setting $\Gamma = 1$ gives the expected maintenance cost under worst-case realizations of uncertainty; this means that the predictive maintenance strategy becomes more conservative to reduce the risk of costs from a too-late maintenance intervention. Conversely, setting $\Gamma = 0 $ results in cost estimates that are more sensitive to the average prediction $\overline{y}_{p,i}$ and can thus yield lower expected maintenance cost when uncertainty is low. In sum, the $\Gamma$ parameter thus allows decision-makers to tailor the optimization process to their specific operational contexts and risk tolerances.

To solve the above robust optimization problem, we use mixed-integer linear optimization of the CVXPY framework~\citep{cvxpy}. All solver parameters were kept at their default values.

\subsection{Continual machine learning}

Continual machine learning is a general paradigm for updating prediction models as new product domains become available while reducing forgetting on earlier ones. Forgetting occurs when updating a model on a new domain improves performance on the new domain but reduces performance on domains learned earlier. 

For our domain-incremental manufacturing setting, \emph{progressive neural networks} (PNN; \citet{rusu2016progressive}) are especially relevant. PNN retains knowledge by changing the network architecture over time; that is, it freezes the components trained on previous domains and adds a new domain-specific component whenever a new domain is introduced. It is therefore well aligned with our setting, where new product configurations are gradually introduced over time but without sacrificing performance on previously encountered ones. Here, forgetting is reduced by expanding the model architecture as new domains are introduced \citep{rusu2016progressive}. Instead of updating a single set of parameters for all domains, PNN allocates a new small network component for each new product domain while retaining the parameters learned on earlier ones. PNN partitions the model into previously learned parameters that are frozen and trainable domain-specific parameters. Formally, when moving from domain $p-1$ to domain $p$, the model consists of the fixed parameters $\theta_{1:p-1}$ learned from earlier domains and the trainable parameters $\theta^{\mathrm{new}}_p$ for the new domain. Hence, the optimization for learning the network parameters is only over the ``trainable'' parameters $\theta^{\mathrm{new}}_p$. 

The prediction model for domain $p$ can therefore be written as
\begin{equation}
\hat{y}_{p,i}
=
h_p\big(\mathbf{x}_{p,i},\mathbf{m}_{p,i}; [\theta_{1},\dots,\theta_{p-1},\theta^{\mathrm{new}}_p]^T \big),
\end{equation}
where $\theta_{1:p-1}$ remains fixed during training on domain $p$, while only $\theta^{\mathrm{new}}_p$ is updated. This yields the following objective:
\begin{equation}
\underset{\theta^{\mathrm{new}}_p}{\mathrm{arg\,min}}\;
\sum_{i\in\mathcal{I}_p}
\mathcal{L}\big(h_p(\mathbf{x}_{p,i},\mathbf{m}_{p,i}; \theta_{1},\dots,\theta_{p-1},\theta^{\mathrm{new}}_p), y_{p,i}\big).
\end{equation}
This design prevents new training updates from overwriting the parameters learned for earlier domains. In our setting, this is attractive because new product configurations can be learned through domain-specific model components while earlier components remain unchanged. At the same time, PNN does not require replaying historical raw data, which makes it especially relevant when storage constraints are strict. PNN is therefore related to configuration-specific models because it retains domain-specific components, but it differs by organizing them within a single model architecture rather than maintaining fully separate models for each configuration.

\section{Experimental setup}\label{sec:experimental_setup}
To study the operational value of continual learning, we design a series of empirical experiments that compare several strategies for updating the machine learning model.

\subsection{Evaluation protocol}

We design all experiments to closely reflect real-world manufacturing practice at \companyname. While we have access to a large dataset with recorded sensor data for the purpose of benchmarking in this work, this is not the case in real-world scenarios where the volume of generated data prohibits long-term storage. To this end, we simulate the sequential arrival of new product domains using the available dataset, following established practices from the literature \citep[e.g.][]{kirkpatrick2017overcoming,rolnick2019experience,botache2023unraveling}.

To split the data into a train and test set, we enforce a strictly chronological, time-sensitive split for each configuration. Specifically, we order product configurations by their initial production date (see \Cref{tab:dataset_summary}) and, within each configuration, sort all observations chronologically based on their timestamps. We then iterate over the configurations sequentially to reflect the observed sequence of product configurations; this protocol mimics a growing product portfolio in manufacturing, where new configurations enter production over time while earlier configurations remain operationally relevant. The first 80\% of observations form the training period and the subsequent 20\% form the test period, with back-filling applied only within a split. This prevents any look-ahead bias and ensures that evaluation is always conducted on future data relative to the training data. Eventually, all evaluations are reported on held-out test sets with the configurations up to that point in time. 

After training on each product configuration, the underlying machine learning model is updated using continual learning (or one of the other updating strategies as benchmark). Replay-based methods retain a memory buffer of $100$ training samples from earlier configurations, while methods without replay do not store raw observations after the update. Domain adaptation and streaming or drift-aware baselines that require past observations receive the same memory buffer to ensure a comparable data budget. 

\subsection{Operational setting}

Informed by our partner company, we employ the following operational setup. We set the cost of type-I errors to $c_{\mathrm{fp}}=30$~MU (monetary units) and those of type-II errors to $c_{\mathrm{fn}}=50$~MU. The cost of a maintenance event is $c_{{\mathrm{m}}} = 10$~MU. The cost ratio reflects high-quality manufacturing where too-late maintenance is more costly than premature maintenance, as type-II errors may lead to extensive downstream costs through rework. We later also account for settings where the relative cost of preemptive maintenance becomes more costly relative to overdue maintenance. The cost ratio was informed by discussions with our partner company and represents a high-quality manufacturing setting; however, due to confidentiality reasons, all costs are stated in MU.

\subsection{Baselines}

Here, we describe baselines used in our experiments. All methods are integrated into the decision model from \Cref{sec:decision_model} to ensure a fair comparison. Hence, the baselines use the identical robust optimization from Step~2; this way, any performance differences can be attributed only to the learning objective in Step~1.

\subsubsection{Na\"ive baselines}
\begin{enumerate}
    \item \textbf{Static}: This is a na{\"ive} baseline where a neural network is trained once, on the first product configuration, and not updated thereafter; meaning that the parameters remain frozen, and the predictive behavior thus remains static. It serves as a simple comparison to assess the benefit of incorporating newly introduced product domains.
    \item \textbf{Fine-tuning}: This baseline method performs na\"ive sequential training of a neural network on the sequence of product domains. In practice, this means the model is updated using the most recent product configuration, but without any mechanism to preserve the performance on earlier domains.
    \item \textbf{Per-configuration}: This is a baseline where separate neural networks are trained for each configuration. At evaluation time, the prediction is made using the model corresponding to the configuration.
\end{enumerate}

\subsubsection{Domain-adaptation baselines}
We further compare against baselines from domain adaptation, which aim to adapt a model from an old product domain to a new one. These methods are designed to improve performance on the current target domain but, again, do not explicitly preserve performance on earlier domains.

\begin{enumerate}
\item \textbf{Density reweighting (DRW).} DRW reweights observations according to differences between source and target densities \citep{sugiyama2005input}. 

\item \textbf{Nearest-neighbor reweighting (NNRW).} NNRW reweights observations using nearest-neighbor information across source and target domains \citep{loog2012nearest}. 

\end{enumerate}

Both baselines are implemented through the public \texttt{skada} library \citep{gnassounou2024skada}.

\subsubsection{Online learning baselines}

We further compare against online learning methods; these are designed for sequential updating to new domains but these methods have an important caveat in that they do not explicitly aim to preserve performance on earlier domains. 

\begin{enumerate}
\item \textbf{Adaptive Random Forest (ARF).} ARF combines online tree ensembles with drift adaptation and is designed for non-stationary data streams \citep{gomes2017adaptive}. 
\item \textbf{AdaBoost.} AdaBoost is an online boosting method that updates weak learners sequentially as new observations become available \citep{oza2001online}. 
\item \textbf{ADWINBoost.} ADWINBoost augments online boosting with change detection, so that the model can react more quickly to distributional changes \citep{bifet2009new}. 
\end{enumerate}
All baselines for online learning baselines are implemented using the public \texttt{River} library \citep{montiel2021river}. 

\subsubsection{Continual machine learning method}

We focus on a common continual machine learning method that is explicitly designed for settings with domain-incremental learning, namely \textbf{progressive neural networks (PNN).} PNN retains information from earlier domains by expanding the network architecture when a new domain is introduced \citep{rusu2016progressive}. We later also compare PNN against other continual learning methods, namely, EWC \citep{kirkpatrick2017overcoming}, SI \citep{zenke2017continual}, AGEM \citep{chaudhry2018efficient}, and ER \citep{rolnick2019experience}. These comparisons confirm our choice of PNN as the primary continual machine learning method in this paper. We used the public \texttt{mammoth} \citep{buzzega2020dark} library for the implementation.

\subsection{Implementation details}

We instantiate all baselines (where applicable) using neural networks in line with best practice \citep{kraus2020deep}. Specifically, the neural network uses a sequence of four dense layers. The first three layers are followed by a leaky rectifier linear unit activation function. The fourth layer is used as the classification layer, where a sigmoid activation is applied to get probability predictions. 

To ensure fair comparison, we follow best practice and conduct an extensive hyperparameter search on the training data. For this, we use Bayesian optimization via the Optuna framework. Details on the hyperparameter tuning can be found in Online~Appendix \ref{sec:hparam_tuning_details}. Importantly, our hyperparameter tuning procedure is \emph{fair}: all methods share the same hyperparameter search budget of \num{60} trials, which is sufficient for the convergence of the hyperparameter optimization. This ensures that observed performance gains are the result of more effective training objectives rather than differences in tuning effort. 

We set the robustness parameter $\Gamma$, which controls the strength of the prediction uncertainty considered in the robust optimization step as follows. Recall that $\Gamma=0$ means that no uncertainty is taken into account, while, conversely, $\Gamma=1$ is a conservative strategy that considers worst-case prediction. In our experiments, we use $\Gamma=0.5$, which means that during robust optimization at most \SI{50}{\percent} of the uncertainty variables $\pmb{z}$ can be simultaneously $1$. 

For EWC, parameter-importance estimates are computed after training on the currently available configuration data, and only the resulting diagonal importance values together with the corresponding parameter snapshot are retained; no historical raw data or gradient histories are stored. Further implementation details are in Online~Appendix \ref{sec:implementation_details}.

\subsection{Performance metrics}

We measure operational performance using the optimized maintenance cost (in monetary units) defined in \Cref{eq:costs}. Since all methods use the same robust optimization layer, differences in maintenance cost reflect how well the predictions can be translated into cost-effective maintenance decisions. We compute the performance metrics on the configuration-specific held-out test sets, so the results capture performance on previously unseen sensor data and mirror real-world deployment at \companyname. All reported results are averaged over five runs, with standard deviations reported accordingly.

We further report a hypothetical global \emph{lower cost bound}. For this, we train a neural network model jointly on the training data of \emph{all} product configurations. This thus presents a clairvoyant baseline that assumes access to all historical training data (which, however, is not feasible in manufacturing practice due to the high-volume setting). It therefore serves as an empirical lower bound on the achievable maintenance cost. The gap (denoted by ``$\Delta$'') to this lower bound quantifies the cost increase that arises when models must be updated sequentially under data-storage constraints rather than being trained with unrestricted access to all historical data.

We report additional performance metrics for Step~1 (e.g., precision, recall, and AUC) in Online Appendix \Cref{sec:extended_main_results}.

\section{Results}\label{sec:results}
We now empirically compare the operational value of different model-update strategies for predictive maintenance in a domain-incremental high-volume manufacturing setting. We first present the main benchmark results. We then (i)~analyze the cost implications of mitigating forgetting, (ii)~examine maintenance costs at the configuration level, and (iii)~study the sensitivity to different cost ratios. Finally, we also perform a series of robustness checks.

\subsection{Main results}\label{sec:main_results}

We present our main results in \Cref{tab:main_results}. We emphasize that the comparison is fair and explicitly designed to isolate differences in model updating. All methods use the same prediction inputs, the same cost function, and the same robust optimization stage, so differences in maintenance cost arise from how the predictive model is updated over time. All evaluations are out-of-sample, where the chronological order is maintained to prevent information leakage. We rigorously tuned hyperparameters using the same protocol across methods, which ensures that the benchmark compares the update mechanisms rather than differences in implementation choices. Overall, we make three observations:

\begin{table}[htbp]
    \centering
    \footnotesize
    \caption{Main results. Results are averaged over 5 runs $\pm$ standard deviation. \emph{Improvement} is reported over the per-configuration baseline. }\label{tab:main_results}
    
    \begingroup
    \renewcommand{\arraystretch}{1.0}
    
    \begin{tabular}{l
                    S[table-format=7(6)]
                    S[table-format=7]}
    \toprule
    Method  & {Maintenance cost} & {$\Delta$ to lower bound}\\
    \hline\hline
    Lower cost bound  &  113520(6760) & {---}\\
    \hline\hline
    \textbf{Na\"ive baselines} & & \\
    Static &  1039705(394643) & 926185\\
    Fine-tuning & 384310(43006) & 270790\\
    Per-configuration & 261510(679) & 147990\\
    \midrule   
    \textbf{Domain-adaptation baselines} &  & \\
    DRW & 462810(66977) & 349290\\
    NNRW &   261825(955) & 148305\\
    \midrule
    \textbf{Online learning baselines} & &\\
    ARF &  1401375(18095) & 1287855\\
    AdaBoost & 317295(19665) & 203775\\
    ADWINBoost & 496515(129931) & 382995\\
    \midrule
    \textbf{Continual machine learning} & &\\
    PNN & 155890(3861) & 42370\\
    \emph{Improvement} & {\emph{-40.38 \%}} & {\emph{-71.37 \%}} \\
    \bottomrule
    \end{tabular}
    \endgroup
\vspace{-10pt}
\end{table}

First, the decision model instantiated with continual machine learning methods performs best. The continual machine learning method PNN reduces maintenance cost by about {\SI{40}{\percent}} relative to the per-configuration model, and by {\SI{59}{\percent}} relative to fine-tuning. Both the per-configuration model and fine-tuning present strong baselines that reflect industrial practice. PNN also comes close to the hypothetical global lower cost bound, with a remaining gap of only \num{42370} MU. These results demonstrate that continual machine learning yields substantial operational gains in our domain-incremental manufacturing setting. 

Second, the results also confirm the importance of updating the models when new product configurations arrive. Compared with the static model, fine-tuning reduces maintenance cost substantially, and configuration-specific models reduce the cost even further. This shows that models must be adapted to the new product configurations. However, fine-tuning is evidently inferior to the configuration-specific models, which is attributed to the fact that it does not preserve knowledge from earlier product configurations. By contrast, configuration-specific models can avoid some forgetting, but they do not share information across product domains, which explains the superior performance of PNN. Similarly, the domain-adaptation and online learning baselines also incur high maintenance costs, thereby suggesting that these approaches are not sufficient for domain-incremental setting.

Third, the comparison with the strongest non-continual alternatives shows why the structure of PNN is valuable. The per-configuration model and NNRW achieve similar maintenance costs, suggesting that methods tailored to specific configurations can partially address heterogeneity across product domains. However, both approaches remain far from PNN. Per-configuration models do not share information across domains, while NNRW can only reweight a limited set of retained observations. PNN performs better, because it explicitly preserves the learned knowledge for earlier configurations by adding domain-specific parameters for newly introduced configurations. This mechanism is well aligned with our manufacturing setting (cf. \Cref{fig:figure1}), where new product configurations enter over time while earlier configurations may remain operationally relevant or reappear.

\subsection{Cost implications of forgetting}

\begin{figure}[!htbp]
    \centering
    \includegraphics[width=0.4\linewidth]{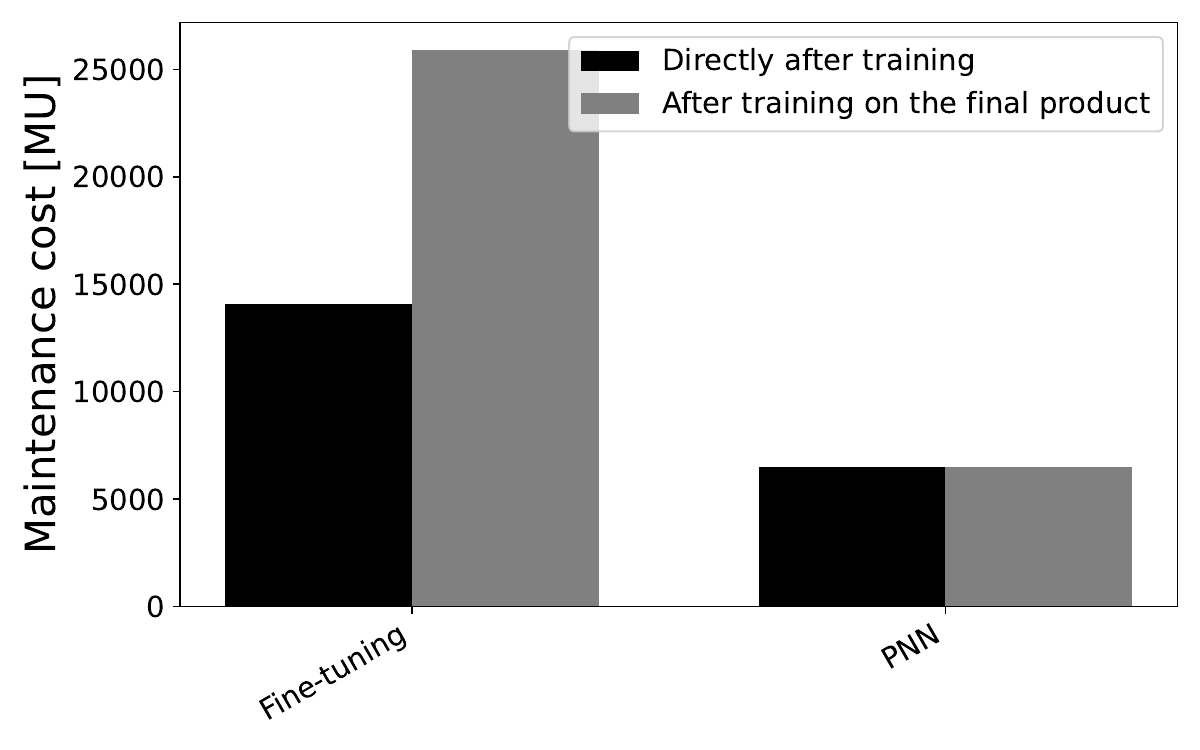}
    \caption{Effect of forgetting on a reoccurring product configuration. Fine-tuning shows a large increase in maintenance cost after subsequent model updates, whereas PNN maintains lower costs over time.}
    \label{fig:operational_benefit_anti_forgetting}
\end{figure}

We now quantify how forgetting affects maintenance costs in our domain-incremental manufacturing setting (see \Cref{fig:operational_benefit_anti_forgetting}). To illustrate this, we focus on product configuration~\#4. Product configuration \#4 is first introduced in the first quarter of year 3 but then reappears at the end of year 3 and year 4 (see \Cref{tab:dataset_summary} from above). For each method, we compute maintenance cost on the held-out test set at two points in the sequence: directly after the model has learned configuration~4 (Years 3 and 4) and after the model has been updated on all subsequent configurations (Q1/Year 5). If forgetting occurs, maintenance costs should increase after the later updates; if a method retains earlier knowledge, costs should remain stable.

The results show a clear pattern. For the fine-tuning baseline, maintenance cost increases considerably after the updates, implying that the model forgets information that was useful for the product configuration. In contrast, PNN keeps maintenance cost nearly unchanged. Together, the results show that forgetting has direct cost implications: in domain-incremental manufacturing, forgetting increases maintenance costs on recurring product configurations, while continual machine learning can avoid such cost increases and maintain low maintenance costs. 

\subsection{Configuration-level cost analysis}

We now analyze performance at the configuration level (see \Cref{tab:per_configuration_analysis_lb_as_reference}). For this, we compare the different methods against the maintenance costs from the lower cost bound. For each of the 25 product configurations, we compute the gap ($\Delta$) in terms of maintenance cost relative to the lower cost bound, where smaller values indicate better performance. The configuration-level analysis allows us to address whether the main result is driven by a small number of product configurations or robust across all product configurations. 

\begin{table}[htbp]
    \centering
    \caption{Gap ($\Delta$) in maintenance cost per product configuration relative to the lower cost bound (lower values are  better). The analysis shows that continual machine learning method PNN generally has the smallest gap.}
    \label{tab:per_configuration_analysis_lb_as_reference}
    \adjustbox{max width=\textwidth}{%
    \begingroup
    \renewcommand{\arraystretch}{1.0}
    \begin{tabular}{lrrrrrrrrrrrrrrrrrrrrrrrrr}
    \toprule
    & \multicolumn{25}{c}{Product configuration \#}\\
    Method & 1 & 2 & 3 & 4 & 5 & 6 & 7 & 8 & 9 & 10 & 11 & 12 & 13 & 14 & 15 & 16 & 17 & 18 & 19 & 20 & 21 & 22 & 23 & 24 & 25 \\
    \midrule
    \textbf{Na\"ive baselines} &&&&&&&&&&&&&&&&&&&&&&&\\
    Static & $3160 \pm 1470$ & $28400 \pm 9430$ & $8620 \pm 1840$ & $91940 \pm 1840$ & $127810 \pm 3800$ & $42820 \pm 570$ & $7290 \pm 5300$ & $37180 \pm 12990$ & $16520 \pm 17610$ & $9360 \pm 8440$ & $4080 \pm 4080$ & $13310 \pm 14590$ & $144320 \pm 95170$ & $28660 \pm 11230$ & $47100 \pm 52830$ & $75480 \pm 81760$ & $14160 \pm 7870$ & $17050 \pm 9350$ & $48220 \pm 34470$ & $8190 \pm 6920$ & $83150 \pm 29510$ & $22530 \pm 6550$ & $1370 \pm 210$ & $30260 \pm 3150$ & $15200 \pm 3870$ \\
    Fine-tuning & $1810 \pm 10$ & $6760 \pm 1790$ & $9680 \pm 560$ & $2140 \pm 230$ & $2140 \pm 810$ & $-40 \pm 220$ & $11060 \pm 30$ & $23360 \pm 60$ & $4180 \pm 160$ & $3480 \pm 130$ & $1280 \pm 130$ & $2810 \pm 80$ & $43630 \pm 50080$ & $14000 \pm 5950$ & $8620 \pm 1200$ & $18530 \pm 1190$ & $7760 \pm 40$ & $8780 \pm 1800$ & $19200 \pm 4410$ & $1240 \pm 1370$ & $36320 \pm 13980$ & $12360 \pm 1970$ & $1560 \pm 60$ & $22040 \pm 3080$ & $8080 \pm 90$ \\
    Per-configuration & $1980 \pm 310$ & $1040 \pm 110$ & $-30 \pm 80$ & $2140 \pm 230$ & $2140 \pm 810$ & $-40 \pm 220$ & $11060 \pm 30$ & $23360 \pm 60$ & $4180 \pm 160$ & $3480 \pm 130$ & $1280 \pm 130$ & $2810 \pm 80$ & $-12740 \pm 1390$ & $6980 \pm 210$ & $7990 \pm 1460$ & $18490 \pm 1240$ & $6800 \pm 300$ & $9900 \pm 220$ & $14430 \pm 80$ & $-770 \pm 110$ & $9100 \pm 330$ & $6960 \pm 20$ & $1560 \pm 60$ & $18060 \pm 20$ & $7820 \pm 30$ \\
    \midrule
     \textbf{Domain-adaptation baselines} &&&&&&&&&&&&&&&&&&&&&&&\\   
    DRW & $1440 \pm 370$ & $8500 \pm 2670$ & $9260 \pm 280$ & $2140 \pm 230$ & $2150 \pm 830$ & $-40 \pm 220$ & $11060 \pm 30$ & $23080 \pm 330$ & $4180 \pm 160$ & $3480 \pm 130$ & $1280 \pm 130$ & $5800 \pm 2520$ & $113920 \pm 38710$ & $22000 \pm 4480$ & $11260 \pm 4020$ & $33360 \pm 8340$ & $8820 \pm 1030$ & $5520 \pm 620$ & $25520 \pm 1920$ & $1640 \pm 2030$ & $34060 \pm 10560$ & $8020 \pm 360$ & $1490 \pm 40$ & $8710 \pm 3490$ & $2600 \pm 560$ \\
    NNRW & $1980 \pm 310$ & $1040 \pm 110$ & $-120 \pm 40$ & $2140 \pm 230$ & $2160 \pm 860$ & $-40 \pm 220$ & $11060 \pm 30$ & $24080 \pm 1070$ & $4180 \pm 160$ & $3480 \pm 130$ & $1280 \pm 130$ & $2810 \pm 80$ & $-12740 \pm 1390$ & $6980 \pm 210$ & $8060 \pm 1350$ & $18490 \pm 1240$ & $6800 \pm 300$ & $9900 \pm 220$ & $14430 \pm 80$ & $-770 \pm 110$ & $8760 \pm 90$ & $6880 \pm 110$ & $1560 \pm 60$ & $18060 \pm 20$ & $7820 \pm 30$ \\
    \midrule
    \textbf{Online learning baselines} &&&&&&&&&&&&&&&&&&&&&&&\\
    ARF & $3980 \pm 310$ & $34980 \pm 110$ & $9890 \pm 40$ & $91160 \pm 740$ & $125740 \pm 830$ & $42460 \pm 60$ & $10980 \pm 1240$ & $62200 \pm 9020$ & $31560 \pm 590$ & $16580 \pm 1270$ & $7910 \pm 690$ & $24690 \pm 1670$ & $210700 \pm 1470$ & $36750 \pm 210$ & $94870 \pm 3320$ & $155260 \pm 4460$ & $25790 \pm 1330$ & $28580 \pm 2480$ & $75580 \pm 590$ & $13160 \pm 110$ & $104660 \pm 900$ & $27160 \pm 220$ & $2520 \pm 420$ & $32720 \pm 320$ & $17960 \pm 30$ \\
    AdaBoost & $2060 \pm 250$ & $3300 \pm 870$ & $570 \pm 420$ & $2160 \pm 210$ & $2160 \pm 810$ & $-20 \pm 200$ & $11060 \pm 30$ & $23660 \pm 60$ & $4780 \pm 390$ & $3480 \pm 130$ & $1280 \pm 130$ & $4140 \pm 60$ & $2080 \pm 6290$ & $9830 \pm 1950$ & $10980 \pm 180$ & $27010 \pm 950$ & $7600 \pm 180$ & $10280 \pm 200$ & $    19480 \pm 1760$ & $370 \pm 200$ & $16960 \pm 1610$ & $8960 \pm 810$ & $1590 \pm 10$ & $21020 \pm 1210$ & $8980 \pm 300$ \\
    ADWINBoost & $3600 \pm 1000$ & $15240 \pm 8210$ & $9560 \pm 1250$ & $2180 \pm 160$ & $2480 \pm 370$ & $0 \pm 180$ & $11060 \pm 30$ & $24470 \pm 1130$ & $4200 \pm 140$ & $3480 \pm 130$ & $1280 \pm 130$ & $3720 \pm 870$ & $59950 \pm 48080$ & $18470 \pm 7380$ & $18490 \pm 1130$ & $34780 \pm 10800$ & $7720 \pm 710$ & $11330 \pm 1030$ & $46400 \pm 16100$ & $4140 \pm 4160$ & $48410 \pm 15710$ & $19200 \pm 1290$ & $1580 \pm 40$ & $20380 \pm 1780$ & $10880 \pm 3680$ \\
    \midrule
    \textbf{Continual machine learning} &&&&&&&&&&&&&&&&&&&&&&&\\
    PNN & $-580 \pm 310$ & $840 \pm 150$ & $-120 \pm 40$ & $2940 \pm 490$ & $8040 \pm 3700$ & $900 \pm 290$ & $860 \pm 110$ & $0 \pm 60$ & $500 \pm 180$ & $3500 \pm 1240$ & $100 \pm 220$ & $1540 \pm 130$ & $-11710 \pm 2550$ & $1080 \pm 910$ & $3260 \pm 1280$ & $19500 \pm 3810$ & $490 \pm 550$ & $1410 \pm 160$ & $9140 \pm 520$ & $-760 \pm 130$ & $320 \pm 110$ & $440 \pm 280$ & $700 \pm 250$ & $-20 \pm 320$ & $0 \pm 50$ \\
    \bottomrule
\end{tabular}
\endgroup
}
\end{table}

We make the following observations. First, PNN again performs best. We find that the cost advantages of continual machine learning are not concentrated in only a few configurations. Instead, PNNs achieve small cost gaps for most product configurations, indicating that the benefits are robust across a broad range of product domains. In a few cases, PNN even slightly outperforms the (global) lower cost bound benchmark, indicated by the negative cost gap. This is explained by the fact that the bound is empirically estimated by training jointly on all configurations and subject to finite-sample variation. Overall, this analysis reinforces the main result: the lower maintenance costs from continual machine learning are not driven by isolated configurations, but reflect a robust pattern across the product portfolio.

\subsection{Sensitivity to different cost ratios}\label{sec:different_cost_ratio}

We now repeat the main experiment under a different cost ratio (see \Cref{tab:cost_ablation}). Here, we set the ratio of type-I to type-II errors as $\frac{c_{\mathrm{fp}}}{c_{\mathrm{fn}}} = \frac{30}{20}$; that is, we account for settings where the cost of preemptive maintenance becomes more costly relative to overdue maintenance. These changes affect only the robust optimization in Step~2; Step~1 predictions remain unchanged.

\begin{table}[htbp]
\centering
\small
\caption{
Results for a different cost setting of $\frac{c_{\mathrm{fp}}}{c_{\mathrm{fn}}} = \frac{30}{20}$, which makes the cost of preemptive maintenance more costly relative to overdue maintenance. Results are averaged over 5 runs $\pm$ standard deviation. \emph{Improvement} is reported over the per-configuration baseline.}\label{tab:cost_ablation}
\begingroup
    \renewcommand{\arraystretch}{1.0}
\begin{tabular}{     l
                     S[table-format=7(6)]
                     S[table-format=7]
                     }
    \toprule
    Method &  {Maintenance cost} & {$\Delta$ to lower bound}\\ 
    \hline\hline
    Lower cost bound  &  83555(4151) & {---}\\
    \hline\hline
    \textbf{Na\"ive baselines} & & \\
    Static  & 960115(503552) & 876560\\
    Fine-tuning & 229210(44321) & 145655\\
    Per-configuration & 104910(679) & 21355\\
    \midrule
    \textbf{Domain-adaptation baselines}  & &\\
    DRW & 341850(89548) & 258295\\
    NNRW & 105225(955) & 21670\\
    \midrule
    \textbf{Online learning baselines} & & \\
    ARF &  1323930(16801) & 1240375\\
    AdaBoost & 160695(19665) & 77140\\
    ADWINBoost &  339915(129931) & 256360\\
    \midrule
    \textbf{Continual machine learning} &  & \\
    PNN & 89955(1648) & 6 400\\
    \emph{Improvement} & {\emph{-14.26 \%}} & {\emph{-70.04 \%}} \\
    \bottomrule
    \end{tabular}
    \endgroup
\vspace{-10pt}
\end{table}

The results are in line with our previous findings (see \Cref{tab:main_results}). We first note that, in line with our expectations from the imbalance in our manufacturing dataset, the overall maintenance costs are lower for a cost ratio that makes type-I errors more costly relative to type-II errors. Again, PNN outperforms the baseline. Importantly, PNN comes closest towards the theoretically achievable lower cost bound. Together, the results for the new cost settings again demonstrate a large operational value of using continual machine learning for predictive maintenance in domain-incremental settings.

\subsection{Robustness checks}

\subsubsection{Robustness check using other continual machine learning methods}\label{sec:robustness_check_cl_methods}

We now compare PNN against several additional continual machine learning approaches, namely, EWC \citep{kirkpatrick2017overcoming}, SI \citep{zenke2017continual}, AGEM \citep{chaudhry2018efficient}, and ER \citep{rolnick2019experience}. The results are reported in \Cref{tab:other_cl_methods}. We observe that PNN performs best overall by a substantial margin. The reason is that the other approaches mainly reduce forgetting indirectly, for example by changing the loss function or constraining parameter updates through regularization. This can help preserve earlier knowledge, but it does not explicitly account for the heterogeneity across product configurations. PNN, in contrast, gives each product configuration its own model component, which allows the model to capture configuration-specific patterns learned earlier while adding new components for newly introduced configurations. The results thus confirm that PNN is well-aligned with our domain-incremental manufacturing setting. 

\begin{table}[htbp]
    \centering
    \small
    \caption{Comparison of PNN against other continual machine learning methods. Results are averaged over 5 runs $\pm$ standard deviation. }\label{tab:other_cl_methods}
    \begingroup
    \renewcommand{\arraystretch}{1.0}
    \begin{tabular}{l
                    S[table-format=7(6)]
                    S[table-format=6]
                    }
    \toprule 
    Method  & {Maintenance cost} & {$\Delta$ to lower bound} \\
    \hline\hline
    Lower cost bound  &  113520(6760) & {---}\\
    \hline\hline
    EWC  & 264405(3118) & 150885\\
    SI  & 263800(4016) & 150280\\
    AGEM & 261300(127) & 147780\\
    ER & 244570(9645) & 131050\\
    PNN & 155890(3861) & 42370\\
    \bottomrule
    \end{tabular}
    \endgroup
\vspace{-10pt}
\end{table}

\subsubsection{Robustness check using meta-information}\label{sec:meta_information}

We now repeat our main experiment and include the meta-information (see \Cref{tab:meta_information} in the Online Appendix \Cref{sec:extended_robustness_results}). This is motivated by manufacturing practice, where meta-information captures process characteristics that are not directly recorded by sensors and are often not used in predictive models. Such meta-information may help models distinguish the heterogeneity among product configurations, and therefore provides a stricter test of whether PNN still adds value through its configuration-specific model components. We thus add the meta-information to the sensor data as additional input features. Our observations are consistent with our previous findings: PNN achieves a lower maintenance cost than the baselines and has the smallest gap to the lower cost bound. 

\subsubsection{Robustness across different training data sizes}\label{sec:sensitivity_ds_sizes}

In the Online Appendix \Cref{sec:dataset_sizes_ablation}, we perform an additional sensitivity analysis where we vary the size of the training data. This allows us to test whether the relative performance of the methods depends on the amount of data available for model updates. Again, the results are consistent with our previous findings: continual machine learning outperforms baseline methods by a clear margin and leads to lower maintenance costs.

\section{Discussion}\label{sec:discussion}

\subsection{Implications for decision analytics}


ML/AI are increasingly used to support data-driven decision making across organizational settings, including manufacturing \citep[e.g.,][]{march2019predictive,ferreira2022remaining,scaife2024improve}. Here, an important application is predictive maintenance, where models use sensor data to assess machine health and support proactive maintenance decisions before failures occur \citep{dinh2024predictive}. However, as product diversity grows over time, organizations face a domain-incremental learning problem: prediction models must be updated for newly introduced product configurations while maintaining performance on previous configurations that remain relevant. This challenge is especially pronounced in high-volume manufacturing, where the volume of machine data often makes long-term data retention infeasible. Our study assesses the operational value of continual machine learning for updating predictive maintenance models as new product configurations are added over time.

Our study shifts attention to an overlooked challenge in machine-learning-based decision support: updating predictive-maintenance models under domain-incremental learning in high-volume manufacturing. In this setting, models must learn from newly introduced product domains without overwriting knowledge from earlier domains, while the full history of raw sensor data cannot be stored. This differs from much of the existing literature, which often assumes that models can be retrained when needed \citep{cui2018operational,bastani2021predicting} or focuses on adapting to a current target domain without maintaining a high performance on earlier, still-relevant domains \citep{senoner2023addressing}. In contrast, our application at \companyname combines both requirements: prediction models must adapt to newly introduced product configurations while preserving performance on earlier configurations that remain in production or reappear later.

To study the operational value of continual machine learning, we perform a rigorous empirical comparison where we integrate different model update strategies into a tailored maintenance decision model. In the first step, each strategy updates a prediction model sequentially as new product domains become available, under the constraint that the full history of raw sensor data cannot be retained. In the second step, the resulting predictions are translated into maintenance decisions through a predict-then-optimize framework \citep[e.g.,][]{bastani2022applied,elmachtoub2022smart,senoner2023addressing}. We further incorporate robust optimization to account for predictive uncertainty, which is important because maintenance needs are rare but costly when missed. This setup allows us to compare update strategies not only by predictive accuracy, but by their downstream maintenance costs under the same operational objective.

Taken together, the appropriate model update strategy depends on how product domains evolve over time; see also \Cref{sec:extended_related_work} in the Online Appendix. Domain adaptation is more suitable for one-time or limited source-to-target transfers, such as adapting prediction models to a new product, customer segment, or engineered production setting, while performance on the earlier domain is no longer central \citep[e.g.,][]{bastani2021predicting,ng2023augmenting,senoner2023addressing}. Online learning is designed for when the data-generating process changes continuously and the main objective is strong performance on the current distribution, as in demand forecasting applications \citep[e.g.,][]{zenisek2019machine,chen2023network,ferreira2023demand}. In contrast, we find that continual machine learning is more suitable for domain-incremental settings such as ours, where new product domains are added over time while earlier domains remain operationally relevant or reappear, and where storing the full history of raw data is infeasible.

\subsection{Managerial implications}

With the rise of the  Internet of Things and trends such as Industry 4.0, manufacturing equipment is increasingly equipped with sensors for data recording \citep{feng2018research}. In industrial setups, it is common that machines produce several terabytes of data per day. While the generated data could, in principle, offer large potential for in-depth analyses of machine states, the data are generally too large and too costly to be stored long-term, and are typically only retained temporarily, if at all. As a result, managers face a dilemma: predictive maintenance models must be updated as new product configurations and operating conditions emerge, but these updates should not come at the expense of a performance loss on earlier product domains that remain operationally relevant. Thus, managers must seek strategies that can incorporate newly available manufacturing data without relying on full access to historical data for retraining. 

Beyond immediate cost savings, our findings are relevant for organizations seeking to scale machine-learning-based decision systems as their operational environments evolve. For the automotive market, an increase in product variety is expected in the coming years, caused by a shift to electric and hybrid forms of mobility solutions. As a result, automotive suppliers anticipate the introduction of many new product configurations across a wide range of component groups, and each variation potentially has different effects on tool wear. Further, automotive suppliers might intend to scale machine learning-based solutions, with the goal of reducing operational expenses through predictive maintenance applications. Together, the expected growth in both product diversity and manufacturing data volumes implies that maintenance models will need to be updated even more frequently in the future. In this context, our work provides actionable guidance.

Based on our findings, we offer three key recommendations for practitioners and managers: \emph{First}, we recommend monitoring the operational data for a potential increase in heterogeneity over time (e.g., using statistical measures such as KL divergence as shown in \Cref{sec:exploratory}). A larger diversity of product configurations, as reflected in increasing divergence scores, can serve as an early indicator that predictive maintenance models need to be updated to cover newly introduced configurations. 

\emph{Second}, managers are especially likely to benefit more from continual machine learning in settings where the product portfolios are expected to grow over time such as in manufacturing. Similar challenges arise in other data-driven decision support settings in which new decision contexts emerge over time while earlier ones remain relevant and historical data may not be fully retained. In healthcare, for example, new patient groups, treatment protocols, hospitals, or diagnostic technologies may enter clinical practice, while models must still perform well for previously observed patient populations \citep{vokinger2021continual}. 
Likewise, in credit risk assessment, new borrower segments, loan products, macroeconomic regimes, or regulatory environments can create new domains, while earlier customer segments remain operationally relevant \citep{wu2022credit}. 

\emph{Third}, our results suggest that managers should adopt continual machine learning early, even when only limited data is available for newly introduced product configurations. A common hesitation in deploying machine learning is the belief that large amounts of new data are required to generate operational value. However, our findings challenge this assumption: limited domain data does not preclude effective model updates. In fact, we demonstrate that, with as few as roughly five days of machine data, predictive maintenance models using continual machine learning already outperform baseline models that cannot adapt to newly introduced product domains (see \Cref{sec:sensitivity_ds_sizes}). Moreover, continual machine learning is relevant even when meta-information about observable product characteristics, such as material types or supplier identifiers, is available. The reason is that, even though meta-information describes differences across product configurations, continual machine learning can additionally preserve configuration-specific predictive patterns.

\subsection{Concluding remarks}

Predictive maintenance allows for proactive machine repairs before failures occur, but the implementation is challenging in domain-incremental learning manufacturing settings, where product portfolios grow over time, newly introduced products must be incorporated into prediction models, earlier products may remain relevant, and machine data are often too large to store. Using real-world data from \companyname, we empirically compare different model-update strategies in such a domain-incremental manufacturing setting. Through extensive experiments, we study the operational value of continual machine learning. Our results provide guidance on how to update machine-learning-based decision systems when new domains emerge over time, earlier domains remain relevant, and historical data cannot be fully retained.


\setstretch{2.0}
\bibliographystyle{spbasic-bise}
\bibliography{refs_cleaned}


\pagebreak
\section*{Predictive maintenance in manufacturing: The operational value of continual machine learning}




\subsection*{Appendix}

\setcounter{section}{0}
\renewcommand{\thetable}{A\arabic{table}}
\renewcommand{\thefigure}{A\arabic{figure}}
\renewcommand{\thesection}{\Alph{section}}

\section{Extended related work}\label{sec:extended_related_work}
We below discuss the related work in more detail.

\subsection{Domain adaptation}

Domain adaptation studies how a model trained on a \emph{source} domain can be transferred to a different but related \emph{target} domain when the underlying data distributions differ \citep{pan2009survey,kouw2018introduction,kouw2019review,wilson2020survey}. Let product domain $p \in P$ denote a source configuration and let product domain $q \in P$ denote a target configuration observed later in time. Using the notation from \Cref{sec:decision_model}, a source-domain dataset can be written as $\mathcal{D}_p=\{(\mathbf{x}_{p,i},\mathbf{m}_{p,i},y_{p,i})\}_{i\in\mathcal{I}_p}$, whereas the target-domain data are $\mathcal{D}_q=\{(\mathbf{x}_{q,j},\mathbf{m}_{q,j},y_{q,j})\}_{j\in\mathcal{I}_q}$. The objective is to learn a predictor that achieves low error on the target domain despite the fact that $\mathbb{P}_p(X,M,Y)\neq \mathbb{P}_q(X,M,Y)$.

A standard perspective is that the target-domain risk can be bounded by three components: source-domain risk, a divergence term that measures the discrepancy between source and target distributions, and an irreducible term reflecting how well both domains can be fit simultaneously \citep{ben2010theory}. In practice, many domain-adaptation methods therefore optimize a source-loss term together with a domain-alignment term; e.g.,
\begin{equation}
\min_{\theta}\;
\sum_{i\in\mathcal{I}_p}L\!\left(h_{\theta}(\mathbf{x}_{p,i},\mathbf{m}_{p,i}),y_{p,i}\right)
+\lambda\,\Omega(\mathcal{D}_p,\mathcal{D}_q),
\end{equation}
where $L(\cdot)$ is the prediction loss and $\Omega(\mathcal{D}_p,\mathcal{D}_q)$ penalizes differences between source and target domains. Existing methods differ mainly in how they define this transfer mechanism. Sample-based approaches reweight source observations to better match the target domain \citep{pan2009survey,loog2012nearest,kouw2019review}. Feature-based approaches learn representations that are more similar across domains, for example through moment matching or adversarial alignment \citep{ganin2015unsupervised,wilson2020survey}. Parameter-transfer approaches initialize or constrain target-domain models using parameters learned on the source domain \citep{pan2009survey,kouw2019review}.

Domain adaptation has also been applied in manufacturing settings in which operating conditions change across machines, processes, or cutting parameters. Examples include domain-adversarial transfer for generalized tool-wear prediction and related cross-condition wear-detection settings \citep{wang2020domain,kubik2023deep}. These studies show that domain adaptation can be useful when a model must be transferred from one known operating regime to another and the primary goal is performance on the current target condition.

At the same time, domain adaptation has an important caveat for our setting. Domain-adaptation methods typically assume a one-off source-to-target transfer and are evaluated mainly by the target-domain performance \citep{ben2010theory,kouw2019review,wilson2020survey}. In contrast, our manufacturing setting is a domain-incremental learning setting: new product domains are added sequentially, earlier domains remain operationally relevant or reappear, and historical data cannot be fully retained. Under these conditions, the challenge is not only to adapt to the current target domain, but also to preserve performance across previously encountered domains. 

\subsection{Online learning}

Online learning (sometimes also referred to as streaming learning) studies settings in which observations arrive sequentially and models must be updated under memory and computational constraints \citep{hazan2016introduction,gama2014survey}. In these settings, the learner typically observes a sequence of examples
\begin{equation}
\{(\mathbf{x}_t,\mathbf{m}_t,y_t)\}_{t=1}^{T} .
\end{equation}
At each time step $t$, the current model $h_t$ predicts the outcome from $(\mathbf{x}_t,\mathbf{m}_t)$, one then computes the loss $\ell(h_t(\mathbf{x}_t,\mathbf{m}_t),y_t)$ on the observation, and finally updates before the next example arrives. A common objective is to minimize cumulative loss (or equivalently regret relative to a reference predictor), i.e.,
\begin{equation}
\mathrm{Regret}(T)
=
\sum_{t=1}^{T}\ell(h_t(\mathbf{x}_t,\mathbf{m}_t),y_t)
-
\min_{h \in \mathcal{H}}
\sum_{t=1}^{T}\ell(h(\mathbf{x}_t,\mathbf{m}_t),y_t).
\end{equation}
Online learning is especially useful in high-volume applications where storing the full history is infeasible \citep{hazan2016introduction} and where, therefore, only one pass over the data is possible.

Closely related is drift-aware learning (also called concept-drift learning), which considers settings in which the data-generating process changes over time, that is,
\begin{equation}
\mathbb{P}_t(X,M,Y) \neq \mathbb{P}_{t'}(X,M,Y)
\quad \text{for some } t \neq t'.
\end{equation}
The concept-drift literature distinguishes between several adaptation mechanisms, including sliding windows, change detectors, adaptive ensembles, and streaming trees \citep{gama2014survey}. Sliding-window methods focus learning on the most recent observations, which essentially discounts older data when the environment changes. Drift detectors such as ADWIN adapt the effective training window when statistically significant changes are detected \citep{gama2014survey,bifet2009new}. Streaming tree methods such as VFDT and extensions incrementally update tree structures using one-pass statistics \citep{hulten2001mining}. Ensemble-based methods such as online bagging and boosting, as well as Adaptive Random Forest, combine incremental base learners and often include explicit drift-adaptation mechanisms \citep{oza2001online,gomes2017adaptive}.

These methods are well suited when the main objective is rapid adaptation to the \emph{current} regime under limited storage and computation. In many such settings, previously learned concepts are assumed to become outdated and should therefore be discounted, replaced, or detected only when they recur \citep{gama2014survey,suarez2023survey}. This makes online and drift-aware learning highly relevant for non-stationary environments in which the operational priority is to track the current concept as efficiently as possible.

However, online learning has important caveats for the setting studied in our paper. Our manufacturing setting is not characterized by a single evolving stream in which older concepts simply become obsolete. Instead, product domains are added sequentially, earlier configurations remain operationally relevant or reappear; hence, model updates must preserve performance across this growing portfolio under limited data retention. From this perspective, online learning and drift-aware methods capture the storage and update constraints of the problem well, but they are less well suited to the requirement of retaining performance across still-relevant earlier domains. 

\subsection{Continual machine learning}

The term ``continual machine learning'' refers to a research paradigm for updating machine learning models as new domains are added over time while preserving performance on earlier ones \citep{el2024continual}. A key motivation is to handle distribution shifts that arise when newly observed data differ from the data on which the model was originally trained. In manufacturing, for example, newly introduced product configurations may exhibit different patterns of tool wear, which can make models trained only on earlier configurations less effective. Formally, a distribution shift\footnote{For a comprehensive overview of distribution shifts in machine learning, see \citet{kouw2018introduction}.} occurs when newly observed data $X'$ no longer follow the same distribution as the original training data $X$ \citep{moreno2012unifying,kouw2018introduction}. Consequently, we have $\mathbb{P}(X,Y) \neq \mathbb{P}(X',Y')$, where $Y$ and $Y'$ denote the target variable associated with $X$ and $X'$, respectively.

A na\"ive way to address such shifts is to fine-tune a model on the newly available data \citep{kouw2018introduction,bastani2021predicting}. However, this can overwrite parameters that encode important information learned from earlier domains and thereby reduce performance on previously encountered data. This phenomenon is known as \emph{forgetting} (or \emph{catastrophic forgetting}) \citep{mccloskey1989catastrophic}. Forgetting is particularly relevant in domain-incremental settings, where new domains are added over time while earlier ones remain operationally relevant. 

To address this challenge, the continual machine learning literature has developed a range of model-update strategies, which can be grouped into three broad categories \citep{wang2024comprehensive}: (i)~replay-based, (ii)~architecture-based, and (iii)~regularization-based approaches. \emph{Replay-based} approaches mitigate forgetting by retaining data from earlier domains in memory \citep{rebuffi2017icarl,rolnick2019experience} or by generating synthetic samples using generative models \citep{mundt2022unified}.  \emph{Architecture-based} approaches modify the network architecture to accommodate new domains \citep{rusu2016progressive,qin2021bns}. These methods typically expand the model by adding new components for newly observed domains while keeping older components fixed. \emph{Regularization-based} approaches mitigate forgetting by penalizing large changes in model parameters when learning from a new domain \citep{zenke2017continual,kirkpatrick2017overcoming}. These methods assume that the current parameter values encode knowledge that should be preserved, so large deviations may lead to a loss of previously acquired information.

So far, continual machine learning has been studied primarily on image datasets \citep[cf.][]{kirkpatrick2017overcoming,zenke2017continual,janetzky2024slowing,saha2021gradient}. In contrast, applications to manufacturing remain scarce and often rely on simulated rather than real-world data \citep{maschler2022regularization,tercan2022continual}. To the best of our knowledge, \citet{maschler2021regularization} is the only study that applies continual machine learning in manufacturing, but for anomaly detection in battery production rather than predictive maintenance. Taken together, the existing literature provides only limited evidence from real-world manufacturing settings in which product domains are added sequentially, earlier configurations remain relevant or reappear, and historical data cannot be fully retained. We therefore empirically study the operational value of continual machine learning methods in high-volume manufacturing.

\newpage
\section{Implementation details}\label{sec:implementation_details}
\textbf{Preprocessing}:
Following best practices \citep{hastie2009elements}, we standardize all features to have zero mean and unit variance. In our implementation, we use PyTorch and \texttt{Python}. The experiments were performed on NVIDIA A100 GPUs provided through \companyname's compute cluster.

\textbf{Undersampling}. In our experiments, we address the imbalance by incorporating re-sampling during training. Concretely, we apply under-sampling of the majority (“no-maintenance”) class within training, while keeping the evaluation distribution unchanged to reflect the real-world base rate. We chose under-sampling as the primary remedy because it is simple to reproduce and integrates cleanly across all baselines. However, alternative methods for dealing with class imbalances such as weighted losses are also compatible with our pipeline.

\textbf{Empirical Fisher computation.}
To compute the Fisher, we proceed as follows. After finishing training on a product configuration, we compute the diagonal empirical Fisher using gradients on the currently available training data for that segment. Here, we use the gradients that are available for the current training data. Then, we store per-parameter importance values from the diagonal Fisher, but neither historical samples nor full gradient histories. When moving to the next configuration, we update the stored importance estimates with the Fisher estimate from the newly observed configuration, following established practice. See \citet{kirkpatrick2017overcoming,aich2021elastic} for further details on the Fisher computation.

\newpage
\section{Hyperparameter tuning}\label{sec:hparam_tuning_details}
\Cref{tab:hparam_tuning} lists the tuning ranges for all hyperparameters. We implemented the tuning using the \texttt{optuna} library and performed Bayesian optimization over \num{60} trials. For each approach, we used the same number of trials, providing each method with an equal computational budget to ensure a \emph{fair} comparison. All neural network baselines share the same search space, and for continual machine learning methods, we included method-specific parameters.

\begin{table}[bpht]
    \centering
    \caption{Hyperparameter tuning ranges.}
    \label{tab:hparam_tuning}
    \begin{threeparttable}
    \begin{tabular}{lll}
    \toprule    
       Method & Tuning parameters & Tuning range\\
       \midrule
       Static & Learning rate & 0.00001 to 0.1\\   
       & Batch size & [32, 64, 128, 256] \\

       \midrule
       Fine-tuning & Learning rate & 0.00001 to 0.1\\   
       & Batch size & [32, 64, 128, 256] \\
       \midrule
       DRW & Learning rate & 0.00001 to 0.1\\   
       & Batch size & [32, 64, 128, 256] \\

       & Buffer size & 100\\
       \midrule
       NNRW & Learning rate & 0.00001 to 0.1\\   
       & Batch size & [32, 64, 128, 256] \\

       & Buffer size & 100 \\
       \midrule
       ARF & Bagging $\lambda$ & 1 to 6\\
       & Max features & {sqrt, log2}\\

       \midrule
       AdaBoost{$^\dagger$} &  Learning rate & 0.00001 to 0.1\\

       \midrule
       ADWINBoost{$^\dagger$} &  Learning rate & 0.00001 to 0.1\\

       \midrule
       EWC & $\lambda_{\mathrm{F}}$ & 0 to 1000\\
       & Learning rate & 0.0001 to 0.1\\

       \midrule
       SI & $\lambda_{{\mathrm{SI}}}$ & 0 to 1000\\
       & Learning rate & 0.0001 to 0.1\\
       & Batch size & [32, 64, 128, 256] \\

       \midrule
       AGEM & Learning rate & 0.0001 to 0.1\\
       & Batch size & [32, 64, 128, 256] \\

       & Buffer size & 100\\
       \midrule
       ER & Learning rate & 0.0001 to 0.1\\
       & Batch size & [32, 64, 128, 256] \\

       & Buffer size & 100\\
       \midrule
       PNN & Learning rate & 0.0001 to 0.1\\
       & Batch size & [32, 64, 128, 256] \\

       \midrule
       Lower cost bound$^\ddagger$ & Learning rate & 0.0001 to 0.1\\
       & Batch size & [32, 64, 128, 256] \\

       \bottomrule
    \end{tabular}
    \begin{tablenotes}[flushleft]
        \scriptsize
        \item[$\dagger$] Learning rate used for the boosting model.
        \item[$\ddagger$] Model trained on all configurations jointly.
    \end{tablenotes}
    \end{threeparttable}
\end{table}

\newpage
\section{Extended main results}\label{sec:extended_main_results}
In \Cref{tab:extended_main_results}, we report F1, precision, recall, and area-under-curve (AUC) scores for the main results. These results provide additional support for our findings; and methods with stronger predictive performance generally also achieve lower maintenance cost in the main paper. In particular, PNN shows the strongest predictive performance among the feasible sequential methods, and comes closest to the lower cost bound across all reported metrics.

At the same time, maintenance costs depend not only on correct predictions, but also on which type of error is made and how the prediction is translated into a maintenance decision by the decision layer. For example, false negatives may be more costly than false positives when delayed maintenance causes downstream defects or rework. Moreover, we use ensemble predictions and a robust optimization step, so both the mean prediction and the predictive uncertainty affect the final maintenance decision. Since the reported scores are averaged across product configurations and repeated runs, they can well capture the overall predictive trend, but may mask configuration-level differences. Such differences matter operationally, because weak performance on a recurring or high-risk configuration can lead to disproportionately high maintenance cost. For this reason, we also conduct configuration-level analyses and evaluate methods through actual downstream maintenance cost.

\begin{table}[htbp]
    \centering
    \caption{Extended results.}
    \label{tab:extended_main_results}
    \footnotesize
    \begingroup
    \renewcommand{\arraystretch}{1.1}
    \begin{tabular}{lS[table-format=2.2(4)]S[table-format=2.2(4)]S[table-format=2.2(4)]S[table-format=2.2(4)]}
    \toprule
    Method & {F1} & {Precision} & {Recall} & {AUC} \\
    \midrule
    Lower cost bound & 0.92(0.10) & 0.91(0.11) & 0.95(0.06) & 0.99(0.02) \\
    \midrule
    \textbf{Na\"ive baselines} & & & &\\
    Static & 0.53(0.09) & 0.58(0.13) & 0.56(0.09) & 0.59(0.18) \\
    Fine-tuning & 0.61(0.15) & 0.63(0.16) & 0.64(0.15) & 0.71(0.18) \\
    Per-configuration & 0.75(0.15) & 0.75(0.16) & 0.79(0.14) & 0.90(0.10)\\
    \midrule
    \textbf{Domain-adaptation baselines} &&&&\\
    DRW & 0.65(0.16) &  0.68(0.17)  &0.69(0.16) & 0.78(0.18) \\
    NNRW & 0.53(0.07) &  0.58(0.12) & 0.56(0.08) &  0.61(0.16) \\
    \midrule
    \textbf{Online learning baselines} &&&&\\
    ARF &  0.53(0.09) & 0.56(0.12) & 0.57(0.11) & 0.57(0.12) \\
    AdaBoost & 0.58(0.21) & 0.67(0.13) & 0.71(0.12) & 0.71(0.12) \\
    ADWINBoost & 0.58(0.18) & 0.64(0.15) & 0.68(0.13) & 0.70(0.14) \\
    \midrule
    \textbf{Continual machine learning} &&&& \\
    EWC & 0.67(0.20) &  0.69(0.20) &  0.71(0.19) & 0.83(0.17) \\
    SI & 0.67(0.21) & 0.70(0.20) & 0.71(0.19) & 0.82(0.19) \\
    AGEM & 0.76(0.16) & 0.76(0.16) & 0.81(0.14) & 0.89(0.12) \\
    ER & 0.81(0.15) & 0.79(0.15) & 0.86(0.12) & 0.93(0.09) \\
    PNN & 0.89(0.12) &  0.88(0.13) & 0.91(0.10) & 0.97(0.04) \\
    \bottomrule
    \end{tabular}
\endgroup
\end{table}

\newpage
\section{Extended robustness results}\label{sec:extended_robustness_results}
In \Cref{tab:meta_information}, we repeat our main experiment and include the meta-information as additional input features. We observe that PNN achieves lower maintenance cost than the baseline methods, and has the smallest gap to the lower cost bound.

\begin{table}[ht]
\centering
\footnotesize
\caption{
Results for including meta-information. Continual machine learning methods outperform the baselines by a considerable margin. We use the default type-I/II error ratio of $\frac{c_{\mathrm{fp}}}{c_{\mathrm{fn}}} = \frac{30}{50}$. Results are averaged over 5 runs $\pm$ standard deviation. Improvement is reported over the per-configuration baseline.
}\label{tab:meta_information}
\begingroup
    \renewcommand{\arraystretch}{1.1}
\begin{tabular}{lS[table-format=6(6)]S[table-format=6]}
\toprule
Method & {Maintenance cost} &  {$\Delta$ to lower bound}\\
\hline\hline
Lower cost bound & 89595(2527) & {---}\\
\hline\hline
\textbf{Na\"ive baselines} && \\
Static & 546770(237208) & 457175\\
Fine-tuning   & 277350(4182) & 187755\\
Per-configuration & 179745(14613) & 90150\\
\midrule
\textbf{Domain-adaptation baselines} & &\\
DRW  & 466307(31452) & 376712\\
NNRW & 287980(11539) & 198385\\
\midrule
\textbf{Online learning baselines} & &\\
ARF  & 872090(186012) & 782495 \\
AdaBoost & 271350(3590) & 181755 \\
ADWINBoost & 535785(140434) & 446190 \\
\midrule
\textbf{Continual machine learning} & &\\

PNN & 117848(8729) & 28253\\
\emph{Improvement} & {\emph{-34.43 \%}} & {\emph{-68.66 \%}} \\
\bottomrule
\end{tabular}
\endgroup
\end{table}

\newpage
\section{Sensitivity to the dataset size}\label{sec:dataset_sizes_ablation}
\begin{figure}[htbp]
    \centering
    \includegraphics[width=0.5\linewidth]{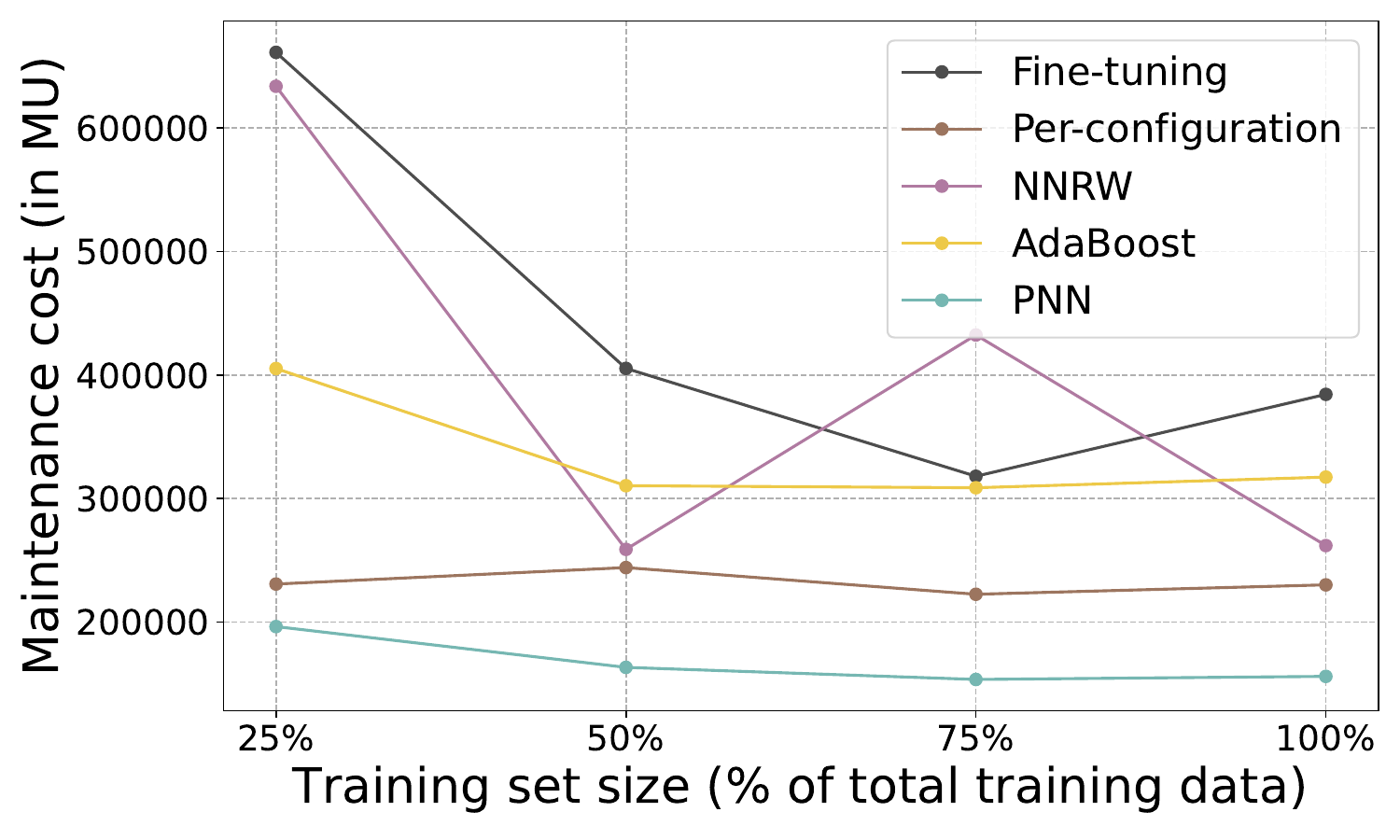}
    \caption{Maintenance cost for different training data sizes. Results show that PNN remain effective across dataset sizes and consistently outperform baseline methods.}
    \label{fig:abl_dataset_size}
\end{figure}

We repeat the main numerical experiment while varying the amount of training data available per product configuration (see \Cref{fig:abl_dataset_size}). This analysis is motivated by the fact that machine learning often requires substantial amounts of training data, while data collection in manufacturing can be costly, especially during the ramp-up phase of deploying new predictive models. By assessing the operational value of different model updating strategies across varying dataset sizes, we provide guidance on the minimum amount of domain data required for these strategies to be effective. Specifically, we vary the amount of available training data by using subsets of size \SI{25}{\percent}, \SI{50}{\percent}, \SI{75}{\percent}, and \SI{100}{\percent} of the total training data for each product configuration. We keep the size of the test set unchanged. For each subset size, we perform an extensive hyperparameter optimization to ensure a fair comparison. We focus on the best-performing methods from \Cref{tab:main_results}, namely: fine-tuning, per-configuration, NNRW, AdaBoost, and PNN.

The results are consistent with our previous findings. Already at \SI{25}{\percent}, corresponding to roughly five days of machine data, the continual machine learning method PNN leads to lower maintenance costs than the baseline methods. From a managerial perspective, this suggests that effective model updates do not require long data-collection periods. Even relatively small amounts of newly available domain data can be sufficient to update predictive maintenance models effectively. Overall, these results indicate that continual machine learning methods remain effective even when the amount of training data per product configuration is reduced. We also observe that maintenance costs generally decrease as the amount of training data increases.

\end{document}